\documentclass[sigconf]{acmart}

\usepackage[english]{babel}
\usepackage{listings}
\usepackage{minted}
\usepackage{amsmath}
\usepackage{booktabs}
\usepackage{graphicx}
\usepackage{hyperref}
\usepackage{subfig}
\usepackage{float}
\usepackage{color,soul}
\restylefloat{figure}

\usepackage{colortbl}
\usepackage{comment}
\usepackage[inline,shortlabels]{enumitem}%
\usepackage{multirow}
\usepackage{tabularx}
\usepackage{url}
\usepackage{xcolor}

\usepackage{array}
\newcolumntype{L}{>{\arraybackslash}m{4cm}}

\newcommand{\metric}[1] {$\mathsf{#1}$}

\newcommand{\myComment}[1]{} 
\newcommand{\our}{$\mathsf{LiCQA}$}
\newcommand{\parab}[1]{\bigskip\noindent\textbf{#1}.}

\begin{document}

\title[A Lightweight Complex Question Answering System]{\our~: A Lightweight Complex Question Answering System}

\author{Sourav Saha}
\affiliation{%
	\institution{Indian Statistical Institute}
	\city{Kolkata}
	\country{India}
}
\email{sourav.saha\_r@isical.ac.in}

\author{Dwaipayan Roy}
\affiliation{%
	\institution{Indian Institute of Science Education and Research}
	\city{Kolkata}
	\country{India}
}
\email{dwaipayan.roy@iiserkol.ac.in}

\author{Mandar Mitra}
\affiliation{%
	\institution{Indian Statistical Institute}
	\city{Kolkata}
	\country{India}
}
\email{mandar@isical.ac.in}

\renewcommand{\shortauthors}{Saha et al.}

\begin{abstract}
Over the last twenty years, significant progress has been made in designing
and implementing Question Answering (QA) systems. However, addressing
\emph{complex} questions, the answers to which are spread across multiple
documents, remains a challenging problem. Recent QA systems that are
designed to handle complex questions work either on the basis of knowledge
graphs, or utilise contemporary neural models that are expensive to train,
in terms of both computational resources and the volume of training data
required. In this paper, we present \our, an unsupervised question
answering model that works primarily on the basis of corpus evidence. We
empirically compare the effectiveness and efficiency of \our\ with two
recently presented QA systems, which are based on different underlying
principles. The results of our experiments show that \our\ significantly
outperforms these two state-of-the-art systems on benchmark data with
noteworthy reduction in latency.


\end{abstract}

\maketitle

\section{Introduction} \label{sec:intro}

Open-domain question answering (QA), the task of providing direct answers
to (factoid) questions, has remained an active area of research within the
Information Retrieval (IR) and Natural Language Processing (NLP)
communities. Most recent work in this area belongs to one of two paradigms:
extracting answers from structured knowledge bases (KBs) or knowledge
graphs (KGs), and extracting answers from unstructured text documents.
While KBs are high-quality sources of organised information, they also have
important limitations: real-life KBs that are up-to-date, and have both
deep and broad coverage remain a holy grail. Much less technical machinery
is required in order to contribute information to, and maintain,
predominantly unstructured textual resources like the Web in general, and
Wikipedia in particular. For the near future, therefore, traditional QA
architectures based on text retrieval and analysis appear to be the
paradigm of choice for flexible, open-domain QA. In this study, we present
one such system, \our. The three notable features of \our\ that together
distinguish it from most other recently proposed QA systems are: (i)~its
unsupervised nature, (ii)~the ability to handle complex questions, and
(iii)~its efficiency. These are discussed in greater detail below.

\parab{Unsupervised nature} An unsupervised QA system attempts to answer a
user-question \emph{Q} given only the question, and a corpus
\emph{C}~\cite{unsupervised-qa}. In contrast, many of the best-known,
modern QA systems are supervised in nature~\citep{factoid_kg_qa, hotpotqa,
  docqa, squad}. 
Given a large number of training samples, each consisting of a question, a
textual passage containing the answer, and the answer itself, such systems
generally `learn' to extract the answer from a given passage \emph{that is
  known to contain the answer}.

An unsupervised QA system like \our\ has some practical advantages over
supervised systems. First, it is concerned with the more general problem of
finding answers to a question from a collection of documents, rather than
finding the precise answer to a question from a piece of text that contains
the answer. Second, it eliminates the need for large volumes of
human-annotated training data. Finally, state-of-the-art methods for
supervised QA systems often employ \emph{end-to-end} Deep Learning
architectures that are computationally expensive to develop, and which need
GPUs for effective deployment. While modules used within the QA pipeline in
\our\ do involve supervised / deep learning techniques for particular
sub-tasks (e.g., a Part of Speech (POS) tagger, trained on text annotated
with POS tags, and sentence embeddings), we bypass the need for the kind of
computational resources required by \emph{end-to-end} supervised QA
systems.


\parab{Handling complex questions} A query is regarded as \emph{complex} if
it involves multiple entities, and multiple relationships between them,
e.g., 

\centerline{\emph{Which \underline{Nolan} films won an \underline{Oscar},
    but missed a \underline{Golden Globe}?}} 

\noindent
In contrast, a query like \emph{Who won the Turing Award in 1970?} would be
regarded as `simple'. Answers to complex questions will often need to be
synthesised by combining evidence from multiple documents. \our\ is
specifically designed to answer such complex queries. None of the labeled
datasets that are commonly used for training and testing the recent
supervised QA systems mentioned above contains an adequate number of
{complex} questions, which comprise our primary focus.

\parab{Lightweight} To the best of our knowledge, QUEST~\citep{quest}, a
recently-proposed system, represents the state-of-the-art for unsupervised,
complex QA systems. For an empirical comparison, therefore, we chose QUEST
as a baseline, along with DrQA~\citep{chen2017reading}. Our experimental
results suggest that \our\ is \textbf{computationally much less expensive}
than the chosen baselines; yet, it achieves significantly better
performance on the datasets used.

\parab{Summary of contributions} In sum, this paper describes a
\emph{lightweight}, \emph{unsupervised} QA system for \emph{complex
  questions}. An empirical comparison with two recent, state-of-the-art
baselines suggests that the proposed system achieves significant
improvements in terms of both effectiveness and efficiency.
While the techniques presented in this report are simple to implement, we
make the source code
available 
to aid reproducibility, comparisons with other
similar systems, and further research\footnote{\url{https://github.com/souravsaha/licqa}}.

\section{Background and Related Work} 
\label{sec:rel-works} 
Early research on large-scale QA generally considered the task of finding
answers to factoid queries from a corpus of unstructured
documents~\citep{Voorhees99thetrec-8}. The traditional strategy established
by these early methods~\citep{singhal:trec8,lasso:trec8,srihari:ie} consisted
of retrieving a set of paragraphs/ documents for a given question, and
subjecting them to more detailed analysis (e.g., by extracting and scoring named entities (NEs) from them). 
In~\cite{lin-reusable-test-collections}, the authors argued that the intent behind document retrieval for QA and traditional searches can be different.
Further, they proposed a synthetic test collection for QA tasks. 

More recently, the structured and interlinked information provided by
Knowledge Graphs and Knowledge Bases has been applied to question
answering~\citep{trec/AhnJMMRS04,lrec/BuscaldiR06,acl/CaiY13,berant-etal-2013-semantic}.
However, the incompleteness of KGs has restricted their use in building
general purpose QA systems. To overcome this limitation, researchers have
successfully used unstructured text documents to supplement KGs for
QA~\citep{sigir/SavenkovA16,acl/FaderZE13,kdd/FaderZE14}.

Most recent work on QA has focused on utilising deep learning models for
the task~\citep{dong:cnn,tan:rnn}. Some of these efforts have also resulted
in the creation of benchmark QA
models~\citep{berant-etal-2013-semantic,UsbeckNHKRN17,chen2017reading,sawant-garg-chak-rama}.
However, these models require significant training data, and
are also computationally expensive.
\emph{DrQA}~\citep{chen2017reading} is one of these state-of-the-art QA systems.
It combines a TF-IDF based document retriever with a multilayered bi-LSTM
network that encodes the paragraphs from the top retrieved documents based
on word embeddings before training two classifiers to identify the starting
and ending locations of the potential answers. 
Nonetheless, it is still distantly supervised on SQuAD, TREC Questions, Web Questions, and Wikimovies dataset.

Although they have proven to be effective for factoid question answering,
most of the QA systems discussed above are designed for questions that have
an answer contained within a single document. These techniques often fail
to retrieve the correct answer for complex questions, where the answers
involve multiple entities spread across several
documents~\citep{www/AbujabalYRW17,cikm/BastH15,
%
%
acl/KhotSC17,bhutani-etal-2020-answering,
naacl/TalmorB18}. 
A method for quantitative evaluation of interactive QA systems was proposed in~\cite{quant-eval-jaist}. The proposed paradigm, called RUTS, incorporates real users, tasks and systems, and unlike the framework presented in this article, RUTS is applicable for the evaluation of user-centered information-access system.

Recently, an unsupervised method \emph{QUEST} has been proposed
by~\cite{quest} for answering complex questions. Unlike the other methods
mentioned above, QUEST does not depend upon any annotated datasets for
training. It finds answers by building and analysing a \emph{quasi}
knowledge graph, constructed from documents retrieved by a search engine in
response to the question. However, the creation of the KG on the fly leads
to significant execution latency in finding the answers to a given
question. 
\section{\our: A lightweight complex question answering model} \label{our_work}

\begin{figure}[t]
    \centering
    \includegraphics[scale=0.34]{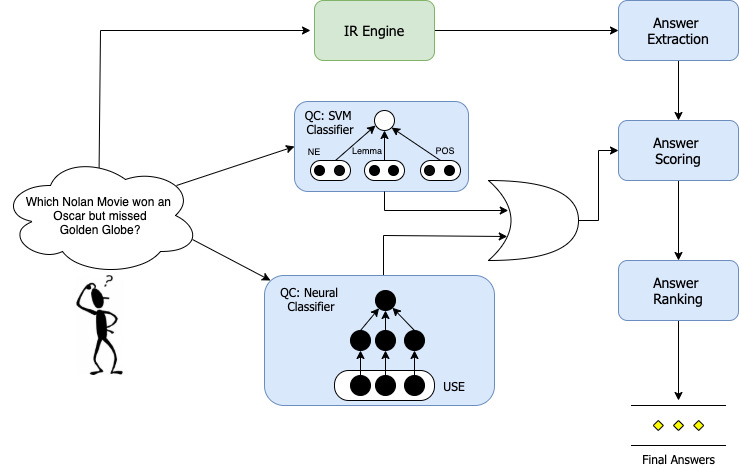}
    \caption{Architecture of the \our\ system.}
    \label{fig:qa}
\end{figure}

\our\ uses a standard pipeline (shown in Figure~\ref{fig:qa}) to process
questions. The question type is determined. Concurrently, a set of
documents that are expected to contain the answer is retrieved using
keywords from the question. Since answers to most questions are entities,
they are extracted from these documents as potential answers. These
entities are then scored, and finally returned in the form of a ranked
list. In the following subsections, each stage of the pipeline is described
in detail.

\subsection{Question Type Classification}
\label{ques_type} 
The question classifier (QC) module tries to classify a given question
according to the predicted type of its answer. For example, consider the
question: \emph{``Which actor played in Troy and Seven?''} Its answer is a
named entity of the \emph{PERSON} type. Similarly, the answer to
\emph{``Where in New Zealand is the Tomb of the Unknown Warrior located?''}
is a \emph{PLACE}-type entity.
QC is generally considered an essential early stage in traditional QA
pipelines~\citep{performance_open_domain_qa}, and has been the focus of
much
research~\citep{qa_type_survey,ques_class_log_linear,high_acc_ques_class,ques_class_grammar}.
For this stage of \our, we experimented with both traditional (`shallow')
and deep learning based methods, discussed below.


\parab{Traditional machine learning based classifier}
\label{sec:traditional-classifiers}
Support Vector Machines (SVMs) are reported to perform well for question
classification~\citep{qa_type_survey}. Following \cite{class_what_ques} and
\cite{trec_ques_classification_data}, we construct the feature vector of a
question $Q$ using: 
\begin{enumerate}[label=(\Roman{*})]
    \item all named entities (NEs) in $Q$, and 
    \item the lemma and POS tag of each word in $Q$.
\end{enumerate}
If all the questions in a test collection
together contain $m$ distinct NEs, $n$ distinct lemmata, $n'$ distinct
lemma-bigrams, $p$ distinct POS tags, and $p'$ distinct POS tag bigrams,
then the feature vector is a binary vector of $m + n + n' + p + p'$
dimensions (see Figure 1).
A linear SVM
  was then trained using the 5500 annotated questions provided by \cite{trec_ques_classification_data}.

\parab{Neural classifiers}
\label{sec:neural-classifiers}
We also try a neural classifier for this stage. A question $Q$ is encoded
employing the 
\emph{Universal Sentence Encoder}
(USE)~\citep{cer-etal-2018-universal}, a state-of-the-art sentence encoder
that is reported to work well for question
classification~\citep{sentence-bert}. This encoder
  embeds
$Q$ into a 256-dimension vector space by passing the average embedding of
word-level unigrams and bigrams present in $Q$ to \emph{DAN}, a deep feed-forward
averaging network. Next, the embedding for $Q$ is fed to a feed-forward
neural network with one dense layer, and a softmax layer at the output
level. The network is trained on the same set of 5500 questions as above,
using the {Adam optimizer}, with {ReLU} as the activation function, and
cross entropy as the loss function.

\subsection{Answer Extraction and Filtering}
\label{sssec:extract_ne}
Given a question $Q$, it is submitted as a query to a standard IR engine
(or search service) to retrieve $\mathcal{D}$, a set of $k$ documents, from
a given collection (or the open Web). Each document $D \in \mathcal{D}$ is
preprocessed: 
accented characters are replaced by their non-accented
equivalents using \texttt{unicodedata} from the Python standard library;
and contractions like \emph{can't} are expanded to
\emph{cannot}.

Next, we extract the set of NEs present in $D$ using
Flair
\citep{flair},
a recent method based on \emph{contextualised} embeddings (where a single
word is mapped to different embeddings, corresponding to the different
sentential contexts in which the word occurs). Each entity is tagged using
a label from the OntoNotes
5 tagset \citep{ontonote5}.

\begin{table}[ht]
  \centering
  \begin{tabular}{l l} 
    \toprule
    Answer Type in~\cite{trec_ques_classification_data} & OntoNotes 5 Entity Type \\ [0.5ex]
    \hline
    HUMAN & PERSON  \\
    LOCATION & GPE, LOC, ORG  \\
    ENTITY & NORP, FAC, PRODUCT, EVENT \\ 
    & LANGUAGE, LAW, WORK\_OF\_ART \\
    NUMERIC & DATE, TIME, PERCENT, MONEY \\
    & QUANTITY, ORDINAL, CARDINAL \\
    money & MONEY \\
    date & DATE \\
    group & ORG \\[0.5ex]
    \bottomrule
  \end{tabular}
  
  \caption{Mapping question / answer types to OntoNotes tags. Answer types
    in uppercase correspond to top-level categories, while 
    lowercases refer to second-level sub-categories.}
  \label{table:entity_mapping_domain}
\end{table}

The final step within this stage involves selecting only entities whose
label (or type) matches the expected answer type for the current question. 
Using Flair for NE extraction poses an operational problem, however. The
tags assigned to candidate answers by Flair and the tags assigned by the QC
module (trained using labels defined by~\cite{trec_ques_classification_data}) are different.
To circumvent this issue, we use a handcrafted table
(Table~\ref{table:entity_mapping_domain}) to map the labels provided
in~\cite{trec_ques_classification_data} to their equivalents in OntoNotes
5. Thus, for a question of type \emph{LOCATION}, all entities labelled
\emph{GPE, LOC} or \emph{ORG} are selected as candidate answers.

\subsection{Answer Scoring}
\label{answer_extract}
The next step involves scoring the selected entities (denoted by
$\mathcal{E} = \{e_1, e_2, \ldots \}$), based on their presumed
relevance to the given question. From the example questions cited above
(i.e., $Q = $ \emph{``Which actor played in Troy and Seven?''}), it is clear
that the answer entity itself is unlikely to occur in $Q$, but contexts in
which the answer occurs and is recognisable as such (i.e., \emph{``Pitt's
  portrayal of Achilles in Troy (2004) $\ldots$ Detective Mills in Seven
  $\ldots$''} etc.) will probably have a substantial overlap. Accordingly, the
key idea in our scoring approach involves measuring the semantic similarity
between $Q$ and the sentences in which a candidate answer occurs.

To keep computational costs down (in keeping with our objective of
designing a system that would be usable in near real-time), we first
compute the \emph{document frequency} (\emph{df}) of each entity $e$ in
$\mathcal{D}$
as follows.
\begin{equation}\label{eq:df}
  \mathit{df}(e) = \lvert \{ D \in \mathcal{D} ~\mid~ e \mbox{ occurs in } D;
  \mbox{ its tag is of the desired type(s)} \} \lvert
\end{equation}
If $\mathcal{E}$ contains more than 100 entities, these entities are ranked
in decreasing order of \emph{df}, and only the top 100\footnote{This number
  was fixed at 100 so as to avoid yet another tunable parameter in our
  system.} are retained for further processing.

Next, we extract from $\mathcal{D}$ all sentences in which any $e \in
\mathcal{E}$ occurs. Let $S_e = \{ s^{(e)}_1, s^{(e)}_2, \ldots, \}$ denote
the sentences in $\mathcal{D}$ that contain $e$, and let $\mathcal{S} =
\bigcup_{e \in \mathcal{E}} S_e$. The match between any $s^{(e)}_i$ and $Q$
is measured using $\cos(\mathbf{Q}, \mathbf{s^{(e)}_i})$, the cosine
similarity between the embeddings of $s^{(e)}_i$ and $Q$. Two types of
embeddings were tried.

\parab{InferSent} In this method proposed by~\cite{infersent}, a pre-trained GloVe embedding of words (300 dimension) are used to get vector representation of the words in the sentences.
Further, a bidirectional LSTM (Bi-LSTM) network is trained on the Stanford Natural Language Inference (SNLI)~\citep{bowman-etal-2015-large} dataset. 
Applying a max pooling over the hidden states of the network, it encodes each sentence to obtain a
4096-dimensional encoded version of the sentence.

\parab{Sentence-Bert} It fine-tunes a pretrained BERT~\citep{bert}/RoBERTa~\citep{roberta}
based on the input sentences, and applies a mean pooling over the output
vectors to create the sentence embedding. As discussed
in~\cite{sentence-bert}, it works significantly faster than BERT and gives
better representation.




\bigskip\noindent Finally, we need to compute a single $\mathit{score}(e)$ for each candidate
answer $e$ by aggregating the scores of all the sentences in $S_e$. Three
aggregation methods were tried.
\begin{itemize}
\item \textbf{Simple averaging:} In this case, $\mathit{score}(e)$,
  denoted by $\mathit{avg\mbox{-}score}(e)$, is given by 
  \begin{equation}\label{eq:avg-scr}
    \mathit{score}(e) = \mathit{avg\mbox{-}score}(e) = \frac{1}{\lvert S_e \lvert}\sum_{s\in S_e} \cos (\mathbf{Q}, \mathbf{s})
  \end{equation}
\item \textbf{Averaging over the best matching sentence from each
    document:} For each document $D \in \mathcal{D}$, we first pick the
  sentence $s_D$ containing $e$ that has the maximum similarity with $Q$
  (see Equation~\ref{eq:sd}).
  \begin{equation}\label{eq:sd}
    s_D = \arg\max_{s \in D \cap S_e} \cos(\mathbf{Q},\mathbf{s})
  \end{equation}
  We then compute the average of these scores over all $D \in \mathcal{D}$.
  \begin{equation}\label{eq:scr_ei}
    \mathit{score}(e) = \mathit{avg\textit{-}max score}(e) = \frac{1}{\lvert \mathcal{D} \lvert} \sum_{D \in \mathcal{D}} \cos(\mathbf{Q},\mathbf{s_d})
  \end{equation}
\item \textbf{Considering only the single best match:} In this case, we
  consider the score of the best matching sentence in $S_e$ as
  $\mathit{score}(e)$.
  \begin{equation}\label{eq:max_score}
    \mathit{score}(e) = \mathit{max\mbox{-}score}(e) = \max_{s\in S_e}\cos (\mathbf{Q}, \mathbf{s})
  \end{equation}
\end{itemize}

\subsection{Answer Ranking}
\label{answer_ranking}
The final ranking of the answers is based on an arithmetic combination of
the similarity score obtained above and the (normalised) document frequency
of the entities. This was motivated by observations from the query
expansion / relevance feedback literature that document frequency is a
useful factor for selecting / weighting expansion terms. 
To combine the two factors, we tried both
weighted addition and simple multiplication.
\begin{itemize}
\item \textbf{Weighted addition:} The semantic similarity score and the
  normalized \textit{df} were combined as follows.
  \begin{equation}\label{eq:mixSimNormDF}
    \mathit{comb}\mbox{-}\mathit{score}_+ = \alpha \times \mathit{score}(e) + \beta \times \frac{\mathit{df}(e)}{\lvert \mathcal{D}\lvert}
  \end{equation}
  Here $\alpha$ and $\beta$ are parameters to be tuned. 
\item \textbf{Simple multiplication:} In this variant, we simply multiplied
  the semantic similarity with the normalized \textit{df}.
  \begin{equation}\label{eq:sim_mul_normDR}
    \mathit{comb}\mbox{-}\mathit{score}_*
    = \mathit{score}(e) \times \frac{\mathit{df}(e)}{\lvert \mathcal{D}\lvert}
  \end{equation}
\end{itemize}
Finally, a list of the entities corresponding to the five highest combined
scores is returned as the set of answers for $Q$. This list is only
\emph{partially} ordered, since \our\ can assign the same rank (one through
five) to \emph{all} entities that obtain the same combined score.

\section{Experimental Setup}
\label{sec:exp}

\subsection{Baselines}
\label{subsec:baseline}
To compare the effectiveness of \our\ with methods having diverse working
principles, we adopt two state-of-the-art complex question answering
models, specifically \textsf{QUEST}~\citep{quest} and
\textsf{DrQA}~\citep{chen2017reading} as our baselines.
Two additional graph-based question answering models,
\textsf{BFS}~\citep{bfs_from_quest} and
\textsf{ShortestPaths}~\citep{quest}, are also considered for comparing the
performance of~\our.

\our\ is designed for answering complex questions (having answers that need
to be constructed from multiple documents) in a real-time application (for
which computational cost 
is an essential factor). Therefore, neither end-to-end QA systems that are
heavily supervised, nor computationally expensive systems based on recent
deep neural models (e.g., BERT~\citep{bert}, RoBERTa~\citep{roberta}) would
be appropriate as baselines for our studies. Note that we do use
\emph{distilled} versions of some of these models in some stages of our
pipeline, however, and compare their effect with that of other alternatives.

\subsection{Datasets}
\label{subsec:dataset}
Datasets that have been commonly used to evaluate QA systems include
SQuAD~\citep{squad}, Natural Questions~\citep{google_nqa} and
WikiMovies~\citep{wikimovies}. As mentioned before, these datasets are not
focused on complex questions, and are designed for supervised QA systems.
In order to experiment with various types of complex questions, and to
allow a fair comparison with our chosen baselines, we use the same dataset as
\textsf{QUEST}. 
Specifically, we use two sets of questions --- WikiAnswers (CQ-W), and
Google Trends (CQ-T) --- for evaluation. Each set contains 150 complex
questions of varying difficulty.

\smallskip \noindent\textbf{Document collection.} Recall from
Figure~\ref{fig:qa} that one of the first steps in \our\ involves
performing a keyword-based retrieval from a corpus with the question to
obtain a set of documents containing the potential answers. In principle,
we treat the open Web as our target document collection, and use a search
service like the Google API to retrieve documents. However, the Web is
dynamic. To ensure reproducibility, the dataset provided by \cite{quest}
includes the 10 top-ranked documents that they actually retrieved in the
course of their experiments for each query. We refer to this set
(comprising the 10 top-ranked documents for each question) as
\textbf{Top10}. Given \textbf{Top10}, we skip the retrieval step during our
experiments, and start by analysing this set of documents.

To minimize the effect of the QA components apparently built into Google's
search engine and to simulate variations in the search algorithm, five
additional document collections were constructed using stratified sampling
from the top $50$ documents returned by Google. These collections are named
\textbf{Strata-1} through \textbf{Strata-5}. These collections also contain
10 documents per query, with $x_1$\% documents selected from within the top
10 ranks, $x_2$\% from documents ranked 11 through 25, and $x_3$\% from
ranks 26 to 50. Table~\ref{tab:strata} shows the values of $x_1$, $x_2$,
and $x_3$ chosen for each these collections.

\begin{table}[h!]
  \centering
  \begin{tabularx}{0.45\linewidth}{cccc}\toprule
    Collection & $x_1$ & $x_2$ & $x_3$ \\\midrule
    \textbf{Top10}    & 100 & 0 & 0  \\
    \textbf{Strata-1} & 60 & 30 & 10 \\
    \textbf{Strata-2} & 50 & 40 & 10 \\
    \textbf{Strata-3} & 50 & 30 & 20 \\
    \textbf{Strata-4} & 40 & 40 & 20 \\
    \textbf{Strata-5} & 40 & 30 & 30 \\\bottomrule
  \end{tabularx}
  \caption{Values of $x_1$, $x_2$ and $x_3$ for the document collections used in this article.}
  \label{tab:strata}
\end{table}

\subsection{Evaluation metrics}
Following standard practice~\citep{quest,convqa_mpi}, we evaluate QA
systems in terms of precision at rank 1 (\metric{P@1}), mean reciprocal
rank (\metric{MRR}), and hit at 5 (\metric{Hit@5}). The commonly used
definitions of these metrics are intended for use with a ranked list that
\emph{does not have any ties}. However, QA systems (including
\textsf{QUEST} and \our) sometimes do produce ties, i.e., multiple answers
that are assigned the \emph{same} rank. In fact, this would be the correct
response to some questions like \emph{Which Nobel Laureate was in the
  Manhattan Project?}) for which multiple correct answers are possible
(\emph{Neils Bohr}, \emph{Enrico Fermi}, etc.). In addition to the above
metrics, therefore, we use tie-aware versions of these
metrics~\citep{najork_ties_paper, tie-aware-jcdl} --- \metric{tP@1}, \metric{tMRR} and
\metric{tHit@5} --- for a more realistic comparison of different systems.
%

\begin{table*}[h!]
\resizebox{0.9\textwidth}{!}{
\begin{tabular}{cccl|rrr}
 \hline
 
            &  Stages of pipeline     & &  & \multicolumn{3}{c}{Evaluation metrics}  \\ [0.5ex] 
\hline
Type classification & Sentence embedding & Scoring & Ranking & tMRR & tP@1 & tHit@5\\
\hline
\multirow{12}{*}{Traditional (SVM)}               & \cellcolor{blue!25}   & \cellcolor{red!25}    & \cellcolor{green!25}\emph{{$\mathit{comb}\mbox{-}\mathit{score}_*$}}  & \textbf{0.412}	& \textbf{0.282}	&  \textbf{0.633}\\ 
                                    & \cellcolor{blue!25}                            & \multirow{-2}{*}{\cellcolor{red!25}max-score}                    & \cellcolor{green!7}\emph{{$\mathit{comb}\mbox{-}\mathit{score}_+$}} & 0.406	 &  0.275  &	0.626\\ 
                                    & \cellcolor{blue!25}                           & \cellcolor{red!15}   & \cellcolor{green!25}\emph{{$\mathit{comb}\mbox{-}\mathit{score}_*$}} & 0.123	& 0.063	& 0.242\\ 
                                    & \cellcolor{blue!25}                            & \multirow{-2}{*}{\cellcolor{red!15}avg-maxscore}                             & \cellcolor{green!8}\emph{{$\mathit{comb}\mbox{-}\mathit{score}_+$}}  & 0.124	& 0.063	& 0.242 \\ 
                                    &  \cellcolor{blue!25}                           & \cellcolor{red!6}   & \cellcolor{green!25}\emph{{$\mathit{comb}\mbox{-}\mathit{score}_*$}} & 0.387	& 0.253	& 0.620 \\ 
                                    & \multirow{-6}{*}{\cellcolor{blue!25}InferSent}                            & \multirow{-2}{*}{\cellcolor{red!6}avg-score}                            & \cellcolor{green!7}\emph{{$\mathit{comb}\mbox{-}\mathit{score}_+$}}  & 0.386	& 0.253	& 0.620\\ 
                                    & \cellcolor{blue!15} & \cellcolor{red!25} & \cellcolor{green!25}\emph{{$\mathit{comb}\mbox{-}\mathit{score}_*$}}   & 0.411	& 0.280	& 0.633\\ 
                                    & \cellcolor{blue!15}                            & \multirow{-2}{*}{\cellcolor{red!25}max-score}                            & \cellcolor{green!7}\emph{{$\mathit{comb}\mbox{-}\mathit{score}_+$}}  & 0.398	& 0.280	& 0.613  \\ 
                                    & \cellcolor{blue!15}                            & \cellcolor{red!15}   & \cellcolor{green!25}\emph{{$\mathit{comb}\mbox{-}\mathit{score}_*$}}  &  0.116	& 0.060	& 0.216\\ 
                                    & \cellcolor{blue!15}                            & \multirow{-2}{*}{\cellcolor{red!15}avg-maxscore}                            & \cellcolor{green!7}\emph{{$\mathit{comb}\mbox{-}\mathit{score}_+$}}   & 0.118	& 0.060	& 0.220\\ 
                                    & \cellcolor{blue!15}                            & \cellcolor{red!5}   & \cellcolor{green!25}\emph{{$\mathit{comb}\mbox{-}\mathit{score}_*$}}   & 0.408 &	0.260	& 0.640\\ 
                                    &  \multirow{-6}{*}{\cellcolor{blue!15}Sent\_BERT}                           & \multirow{-2}{*}{\cellcolor{red!5}avg-score}                            & \cellcolor{green!7}\emph{{$\mathit{comb}\mbox{-}\mathit{score}_+$}}    & 0.393	& 0.260	& 0.613\\ \hline
\multirow{12}{*}{Neural (USE)} & \cellcolor{blue!25}  & \cellcolor{red!25} & \cellcolor{green!25}\emph{{$\mathit{comb}\mbox{-}\mathit{score}_*$}}  & 0.407	& 0.282	& 0.626\\ 
                                    & \cellcolor{blue!25}                            & \multirow{-2}{*}{\cellcolor{red!25}max-score}                            & \cellcolor{green!7}\emph{{$\mathit{comb}\mbox{-}\mathit{score}_+$}} & 0.401	& 0.275	& 0.620\\ 
                                    & \cellcolor{blue!25}                            & \cellcolor{red!15}   & \cellcolor{green!25}\emph{{$\mathit{comb}\mbox{-}\mathit{score}_*$}} &   0.117	 &  0.056	& 0.208\\ 
                                    & \cellcolor{blue!25}                            & \multirow{-2}{*}{\cellcolor{red!15}avg-maxscore}                            & \cellcolor{green!7}\emph{{$\mathit{comb}\mbox{-}\mathit{score}_+$}} &  0.117	&  0.056	& 0.225\\ 
                                    & \cellcolor{blue!25}                            & \cellcolor{red!5}   & \cellcolor{green!25}\emph{{$\mathit{comb}\mbox{-}\mathit{score}_*$}} & 0.385	& 0.260	& 0.613\\ 
                                    & \multirow{-6}{*}{\cellcolor{blue!25}InferSent}                            & \multirow{-2}{*}{\cellcolor{red!5}avg-score}                             & \cellcolor{green!7}\emph{{$\mathit{comb}\mbox{-}\mathit{score}_+$}} &   0.385 	& 0.260	& 0.613\\ 
                                    & \cellcolor{blue!15} & \cellcolor{red!25} & \cellcolor{green!25}\emph{{$\mathit{comb}\mbox{-}\mathit{score}_*$}} &    0.399	&  0.273	&  0.620\\ 
                                    & \cellcolor{blue!15}                            & \multirow{-2}{*}{\cellcolor{red!25}max-score}                            & \cellcolor{green!7}\emph{{$\mathit{comb}\mbox{-}\mathit{score}_+$}}  &  0.389	& 0.273	& 0.600\\ 
                                    & \cellcolor{blue!15}                            & \cellcolor{red!15}   & \cellcolor{green!25}\emph{{$\mathit{comb}\mbox{-}\mathit{score}_*$}}  & 0.106	& 0.050	& 0.226\\ 
                                    & \cellcolor{blue!15}                            & \multirow{-2}{*}{\cellcolor{red!15}avg-maxscore}                            & \cellcolor{green!8}\emph{{$\mathit{comb}\mbox{-}\mathit{score}_+$}}   & 0.104	& 0.050	& 0.206\\ 
                                    & \cellcolor{blue!15}                            & \cellcolor{red!5}   & \cellcolor{green!25}\emph{{$\mathit{comb}\mbox{-}\mathit{score}_*$}}   & 0.389	&  0.246	&  0.620\\ 
                                    & \multirow{-6}{*}{\cellcolor{blue!15}Sent\_BERT}                            & \multirow{-2}{*}{\cellcolor{red!5}avg-score}                            & \cellcolor{green!8}\emph{{$\mathit{comb}\mbox{-}\mathit{score}_+$}}  &  0.387	& 0.260	& 0.600\\ \hline
\end{tabular}
}

\caption{Performance of different stages of pipeline for CQ-W questions in terms of tie aware version of MRR, P@1 and Hit@5.}
\label{tab:my-table}
\end{table*}

\subsection{\our~configuration}
\label{subsec:config}
Other than the parameters involved within the various commodity components
used in our pipeline, the only parameters in \our\ are $\alpha$ and $\beta$
(see Equation~\ref{eq:mixSimNormDF}). These were varied over $[0.1, 0.7]$
in steps of $0.1$. %

\begin{table*}[htb]
\centering
\small
\resizebox{0.999\textwidth}{!}{
\begin{tabular}{l l l l l l l l} 
 \hline
 System         & Metric  & Top10  & Strata-1 & Strata-2 & Strata-3 & Strata-4 & Strata-5 \\ [0.5ex] 
\hline
    
    \textsf{ShortestPaths} & & 0.240 & 0.261 & 0.249 & 0.237 & 0.259 & 0.270 \\ 
    
    \textsf{BFS} &   & 0.249  & 0.256 & 0.266   & 0.212  & 0.219 & 0.254 \\ 
    
    \textsf{DrQA}             & \metric{MRR}                           & 0.226 & 0.237   & 0.257  & 0.256  & 0.215 & 0.248  \\
    
    \textsf{QUEST}            &  & 0.355  & 0.380    & 0.340 & 0.302 & 0.356 & 0.318             \\
    
    \our         & & \textbf{0.432$^*$$^\mathsection$$^{\dagger\ddagger}$} & \textbf{0.424$^*$$^\mathsection$$^{\dagger\ddagger}$} & \textbf{0.431$^*$$^\mathsection$$^{\dagger\ddagger}$} & \textbf{0.420$^*$$^\mathsection$$^{\dagger\ddagger}$} &  \textbf{0.426$^*$$^\mathsection$$^{\dagger\ddagger}$}  & \textbf{0.416$^*$$^\mathsection$$^{\dagger\ddagger}$} 
    \\\midrule
    \% improvement &  & 21.6\% & 11.5\% & 26.7\% & 39\% & 19.6\% & 30.8\%
    
    \\ 
    [0.5ex]\midrule
    
    \textsf{ShortestPaths} & & 0.147 & 0.173 & 0.193 & 0.140 & 0.147 & 0.187                           \\ 
    
    \textsf{BFS} &   &  0.160 & 0.167 & 0.193   & 0.113  & 0.100 & 0.147 \\ 
    
    \textsf{DrQA}             & \metric{P@1}                           & 0.184     & 0.199   & 0.221          & 0.215  & 0.172 & 0.200                           \\
    
    \textsf{QUEST}   &   & 0.268  & \textbf{0.315}  & 0.262 & 0.216 & 0.258 & 0.216  \\
    \our         & & \textbf{0.293$^*$$^{\dagger}$} &  0.280$^\mathsection$ &  \textbf{0.320$^*$$^\mathsection$}  &  \textbf{0.273$^*$$^\mathsection$} & \textbf{0.306$^*$$^\mathsection$$^{\dagger}$}  & \textbf{0.280$^\mathsection$}  \\ \midrule
    \% improvement &  & 9.32\% & -11.1\% & 22.13\% & 26.3\% & 18.6\%  & 29.6\% \\
    %
    [0.5ex]\midrule
    \textsf{ShortestPaths} & & 0.347 & 0.367 & 0.387 & 0.327  & 0.393  &  0.340 \\ 
    
    \textsf{BFS} &   &  0.360 & 0.353 & 0.347    & 0.327   & 0.327  &  0.360 \\ 
    
    \textsf{DrQA}    & \metric{Hit@5}   & 0.313                    & 0.315   & 0.322          & 0.322          & 0.303    &     0.340   \\
    
    \textsf{QUEST}   &   & 0.376 & 0.396 & 0.356 & 0.344 & 0.401 & 0.358  \\
    \our         & & \textbf{0.646$^*$$^\mathsection$$^{\dagger\ddagger}$} & \textbf{0.673$^*$$^\mathsection$$^{\dagger\ddagger}$} & \textbf{0.633$^*$$^\mathsection$$^{\dagger\ddagger}$}  & \textbf{0.653$^*$$^\mathsection$$^{\dagger\ddagger}$} &  \textbf{0.626$^*$$^\mathsection$$^{\dagger\ddagger}$}  & \textbf{0.640$^*$$^\mathsection$$^{\dagger\ddagger}$} \\ 
    [0.5ex]\midrule
    \% improvement &  & 71.8\% & 69.9\% & 77.8\% & 89.8\% & 56.1\% & 78.7\%  \\
\hline
\end{tabular}
}
\caption{Performance comparison of baseline methods with \our~ for CQ-W questions in terms of MRR, P@1 and Hit@5. A superscript $^*$, $^\mathsection$, $\dagger$ and $\ddagger$ respectively indicate that \our~ is statistically significant (t-test with $p < 0.05$) with the \textsf{ShortestPaths}, \textsf{BFS}, \textsf{DrQA} and QUEST. The percentage improvement of \our\ with the best performing baseline (QUEST) is presented in the last row of each group. 
}
\label{baseline_cq_w_t_plus_ourbest}
\end{table*}

\begin{table*}[htb]
\centering
\small
\resizebox{0.999\textwidth}{!}{
\begin{tabular}{ l l l l l l l l} 
 \hline
 System         & Metric  & Top10  & Strata-1 & Strata-2 & Strata-3 & Strata-4 & Strata-5 \\ [0.5ex] 
\hline
    
    \textsf{ShortestPaths} &  & 0.266 & 0.224 & 0.248 & 0.219 & 0.232 & 0.222\\ 
    
    \textsf{BFS} &   & 0.287 & 0.256 & 0.265 & 0.259 & 0.219 & 0.201 \\ 
    
    \textsf{DrQA}             & \metric{MRR}                            & 0.355 & 0.330 & 0.356 & 0.369 & 0.365 & 0.380                           \\
    
    \textsf{QUEST}            &   & \textbf{0.467} & \textbf{0.436} & 0.426 & 0.460 & 0.409 & 0.384             \\
    
    \our         &  & 0.443$^*$$^\mathsection$ & 0.407$^*$$^\mathsection$ & \textbf{0.472$^*$$^\mathsection$$^{\dagger\ddagger}$} & \textbf{0.476$^*$$^\mathsection$$^{\dagger\ddagger}$} & \textbf{0.422$^*$$^\mathsection$} & \textbf{0.468$^*$$^\mathsection$$^{\dagger\ddagger}$}
    \\\midrule
    \% improvement &   & -5.1\% & -6.6\% & 10.8\% & 3.4\% & 3.1\% & 21.8\% 
    
    \\ 
    [0.5ex]\midrule
    
    \textsf{ShortestPaths} &  & 0.190 & 0.140 & 0.160 & 0.160 & 0.150 & 0.130                           \\ 
    
    \textsf{BFS} &    & 0.210 & 0.170 & 0.180 & 0.180 &  0.140 & 0.130 \\ 
    
    \textsf{DrQA}             & \metric{P@1}         & 0.286 & 0.267 & 0.287 & 0.300 & 0.287 & 0.320                           \\
    
    \textsf{QUEST}   &   & \textbf{0.394}  &  \textbf{0.360}   &   \textbf{0.347}   &  \textbf{0.377}  & \textbf{0.333} & 0.288  \\
    \our         &  & 0.306$^*$ & 0.260$^*$ & 0.346$^*$$^\mathsection$ & 0.340$^*$$^\mathsection$ & 0.273$^*$$^\mathsection$ & \textbf{0.320$^*$$^\mathsection$} \\ \midrule
    \% improvement &  & -22.3\% & -27.7\% & -0.2\% & -9.8\% & -18.0\% & 11.1\% \\
    %
    [0.5ex]\midrule
    \textsf{ShortestPaths} &  &  0.350  &  0.320  & 0.340 & 0.310  & 0.330  & 0.290 \\ 
    
    \textsf{BFS} &     &  0.380  & 0.360  & 0.370 & 0.360 & 0.310 & 0.320\\ 
    
    \textsf{DrQA}    & \metric{Hit@5}   & 0.453 & 0.440  &  0.473 & 0.487 & 0.480 & 0.480                         \\
    
    \textsf{QUEST}   &   & 0.531 & 0.496 & 0.510  &  0.500  &    0.503 & 0.459  \\
    \our         &   & \textbf{0.646$^*$$^\mathsection$$^{\dagger\ddagger}$} & \textbf{0.633$^*$$^\mathsection$$^{\dagger\ddagger}$} & \textbf{0.680$^*$$^\mathsection$$^{\dagger\ddagger}$} & \textbf{0.673$^*$$^\mathsection$$^{\dagger\ddagger}$} & \textbf{0.640$^*$$^\mathsection$$^{\dagger\ddagger}$} &  \textbf{0.686$^*$$^\mathsection$$^{\dagger\ddagger}$}\\ 
    [0.5ex]\midrule
    \% improvement &  & 21.6\% & 27.6\% & 33.3\% & 34.6\% & 27.2\% & 49.4\% \\
\hline
\end{tabular}
}
\caption{Performance comparison of baseline methods with \our~ for CQ-T questions in terms of MRR, P@1 and Hit@5. A superscript $^*$, $^\mathsection$, $\dagger$ and $\ddagger$ respectively indicate that \our~ is statistically significant (t-test with $p < 0.05$) with the \textsf{ShortestPaths}, \textsf{BFS}, \textsf{DrQA} and QUEST. The percentage improvement of \our\ with the best performing baseline (QUEST) is presented in the last row of each group. 
}
\label{baseline_cq_t_plus_ourbest}
\end{table*}

\section{Result and Discussions} 
\label{sec:result}
Recall that the \our\ pipeline contains four components for which multiple
options were explored. We first present a comprehensive comparison across
all the different combinations to determine what works best. This also
allows us to carefully analyse the effect of the choice for any individual
module, by fixing the choices for the other modules. In the next
subsection, we present an overall comparison between \our\ and the
baselines mentioned in the previous section. While this does involve
  selecting the most effective design for the pipeline, it \emph{does not}
  amount to retrospectively training parameters on the test data. Finally,
by way of a qualitative comparison with QUEST, as well as failure analysis,
we provide an anecdotal discussion of the results for some specific
questions.

\subsection{Analysis of pipeline components}
\label{sec:analysis-pipline-comp}
To verify the prominence of each component and the robustness of~\our, we vary the different modules and check the performance. 
Table~\ref{tab:my-table} shows the performance of different stages of the pipeline for CQ-W question-set. 
The performance variation with questions from CQ-T has been seen to be similar; hence we are reporting the performances on the CQ-W question-set for the sake of brevity. 
We choose the best module in each stage and report our performance of \our. 

While comparing the overall performance of the question type classifiers, we note that the SVM based classifier works slightly better than the neural classifier.
After extracting the set of sentences from the retrieved documents, the next stage of~\our\ involves scoring of the entities in those sentences which consists of a semantic and a statistical comparison. 
The semantics-based similarity is computed between the question and the extracted sentences containing entities of same type as of the answer. 
Among the two sentence encoding techniques experimented with, \emph{InferSent} has been seen to be working better than \emph{Sent\_BERT} in a majority of cases although the performances are mostly comparable. 
As part of the statistical comparison between the question and the extracted sentences with the NEs, \emph{max-score} (Equation~\ref{eq:max_score}) has been observed to be superior than the other scoring functions (presented in Equation~\ref{eq:avg-scr},~\ref{eq:scr_ei},~\ref{eq:max_score}) irrespective of the other components. 
At the final phase,~\our\ ranks the entities based on the computed similarity scores. 
The best performance is discerned when the normalized document frequency is multiplied with the score of the individual entities using \texttt{$\mathit{comb}\mbox{-}\mathit{score}_*$} (Equation~\ref{eq:sim_mul_normDR}). 
Overall, the optimal performance is reported when the SVM based classifier is used with InferSent as the sentence embedding technique while scoring and ranking of the entities are performed respectively using max-score and $\mathit{comb}\mbox{-}\mathit{score}_*$.
Hence, we report the performance of \our\ based on this configuration and compare the result with the baselines. 

\subsection{End-to-end evaluation}
\label{sec:end-end-evaluation}
\subsubsection{Evaluation using \metric{MRR}, \metric{P@1} and \metric{Hit@5}}
\label{subsec:eval-trad}
Table~\ref{baseline_cq_w_t_plus_ourbest},\ref{baseline_cq_t_plus_ourbest} compare the overall performance
of \our\ and the baselines for both question sets, WikiAnswers (CQ-W) and
Google Trends (CQ-T), in terms
of \metric{MRR}, \metric{P@1} and \metric{Hit@5}.
We observe that \our\ achieved the best performance across all metrics for
CQ-W; it yields an improvement of 21.6\% and 71.8\% over the next best
system (QUEST) in terms of \metric{MRR} and \metric{Hit@5}, respectively,
for the Top10 dataset. Similar trends are observed for Strata 1--5.
The improvements over \textsf{ShortestPaths}, \textsf{BFS} and the strong
baselines, \textsf{QUEST} and DrQA, are statistically significant (paired
t-test with $p < 0.05$) for almost all document sets.

For CQ-T, \our\ achieves the best \metric{Hit@5} across document sets, but
\textsf{QUEST} does consistently better for \metric{P@1}, and also achieves
better \metric{MRR} for the Top10 and Strata-1 sets. It turns out, however,
that the difference between \textsf{QUEST} and \our\ is not statistically
significant in these cases.

Across query and document sets, \our's ability to consistently retrieve at
least one correct answer within the top 5 ranks is particularly
encouraging. In this regard, it is successful for at least 63\% of
the questions, significantly more than the next best system (usually
\textsf{QUEST}).

\begin{table}[h]
  \centering
  Question: \textbf{which country borders mali and ghana?}
  \begin{tabular}{c  l}
    \toprule
    Rank &  Answers   \\ 
    \hline 
    1 & countries $\lvert$ three countries \\
    2 & central bank of west african states $\lvert$ \ensuremath{\ldots} $\lvert$ africa ohada \\
    3 & attack \\
    4 & accra \\
    5 & \textbf{burkina faso} $\lvert$ burkina faso burkina \\\bottomrule
  \end{tabular}
  \caption{Example ranked list returned by QUEST. Rank 2 contains 20 answers. Correct answers are
    in bold face.}
  \label{tab:ties}
\end{table}

\begin{table*}[h!]
\centering
\small
\resizebox{0.999\textwidth}{!}{
\begin{tabular}{l l l l l l l l l l l l l l} 
 \hline
                     &                                & & & \textbf{CQ-W} & & & & & &   \textbf{CQ-T} & & \\ \cmidrule{2-4} \cmidrule{5-8} \cmidrule{9-14}
 System         & Metric  & Top10  & Strata-1 & Strata-2 & Strata-3 & Strata-4 & Strata-5 & Top10  & Strata-1 & Strata-2 & Strata-3 & Strata-4 & Strata-5 \\ [0.5ex] 
\hline

    \textsf{DrQA}             &     &  0.226 &  0.237  & 0.257  & 0.256  & 0.215 &  0.248 &  0.355 &  0.330 &  0.356 & 0.369  &  0.365 &  0.380                           \\
    
    \textsf{QUEST}            & \metric{tMRR}  &  0.067 &  0.116    &  0.066 & 0.067 &  0.077 &  0.055  &  0.065  &  0.075 &  0.066 &  0.076 & 0.069 & 0.057              \\
    \our         & & \textbf{0.412$^{\dagger\ddagger}$} & \textbf{0.411$^{\dagger\ddagger}$} &  \textbf{0.408$^{\dagger\ddagger}$} & \textbf{0.398$^{\dagger\ddagger}$} &  \textbf{0.418$^{\dagger\ddagger}$}  & \textbf{0.399$^{\dagger\ddagger}$} & \textbf{0.433$^{\ddagger}$} & \textbf{0.388$^{\dagger\ddagger}$} & \textbf{0.443$^{\dagger\ddagger}$} &  \textbf{0.455$^{\dagger\ddagger}$}     &  \textbf{0.399$^{\dagger\ddagger}$}  & \textbf{0.434$^{\dagger\ddagger}$} \\ \cmidrule{1-14}
    
    \% improvement & & 82.3\% & 73.4\% & 58.7\% & 55.4\% & 94.4\% & 60.8\% & 21.9\% & 17.5\% & 24.4\% & 23.3\% & 9.31\% & 14.2\% \\[0.5ex]\midrule
    
    \textsf{DrQA}    &  & 0.184 &  0.199 & 0.221  & 0.215  &  0.172 &  0.200    &    0.286 &  \textbf{0.267} &  0.287 &  0.300 & \textbf{0.287} & \textbf{0.320}                           \\
    
    \textsf{QUEST}   &   \metric{tP@1} & 0.040  &  0.093    &   0.043 &  0.042 &  0.053  &  0.034 & 0.043  &  0.042  &  0.036 &  0.048  &  0.035 &  0.031  \\
    \our         & & \textbf{0.282$^{\dagger\ddagger}$} & \textbf{0.270$^{\ddagger}$} & \textbf{0.292$^{\ddagger}$} & \textbf{0.255$^{\ddagger}$} & \textbf{0.303$^{\dagger\ddagger}$} &   \textbf{0.262$^{\ddagger}$} & \textbf{0.296$^{\ddagger}$} & 0.242$^{\ddagger}$ &  \textbf{0.312$^{\ddagger}$} & \textbf{0.320$^{\ddagger}$}      & 0.250$^{\ddagger}$    & 0.290$^{\ddagger}$ \\ \cmidrule{1-14}
    
    \% improvement &  & 53.2\% & 35.6\% & 32.1\% & 18.6\% & 76.1\% & 31.0\% & 3.4\% & -9.3\% & 8.7\% & 6.6\% & -12.8\% & -9.3\% \\[0.5ex]\midrule
    
    \textsf{DrQA}    &    & 0.313 &  0.315  & 0.322 &  0.322 &  0.303   &  0.340  &  0.453  &  0.440  & 0.473  &  0.487 &  0.480 &  0.480   \\
    
    \textsf{QUEST}   & \metric{tHit@5}  & 0.155 &  0.194 &  0.144 &  0.144  &  0.162 &  0.120 &  0.154 &  0.169 &  0.147  &  0.165  &  0.149 &  0.121 \\
    \our         & & \textbf{0.640$^{\dagger\ddagger}$}  & \textbf{0.655$^{\dagger\ddagger}$} &  \textbf{0.614$^{\dagger\ddagger}$}  &  \textbf{0.649$^{\dagger\ddagger}$} &
    \textbf{0.620$^{\dagger\ddagger}$} &  \textbf{0.635$^{\dagger\ddagger}$} & \textbf{0.646$^{\dagger\ddagger}$} & \textbf{0.620$^{\dagger\ddagger}$} &  \textbf{0.657$^{\dagger\ddagger}$} & \textbf{0.662$^{\dagger\ddagger}$}           & \textbf{0.616$^{\dagger\ddagger}$}  & \textbf{0.664$^{\dagger\ddagger}$} \\ [0.5ex]\cmidrule{1-14}
    \% improvement & & 104.4\% & 107.9\% & 90.6\% & 101.5\% & 104.6\% & 86.7\%  & 42.6\% & 40.9\% & 38.9\% & 35.9\% & 28.3\% & 38.3\% \\
\hline
\end{tabular}
}
\caption{Performance comparison of baseline methods with \our~for CQ-W and CQ-T questions in terms of \metric{tMRR}, \metric{tP@1} and \metric{tHit@5}. A $\dagger$ and $\ddagger$ indicate that \our~ is statistically significant (t-test with $p < 0.05$) than \textsf{DrQA} and \textsf{QUEST} respectively. Percentage improvement of \our\ with best performing baseline (DrQA) is presented in last row of each group.}
\label{baseline_cq_w_t_new_metric_plus_ourbest}
\end{table*}


\subsubsection{Evaluation using \metric{tMRR}, \metric{tP@1} and \metric{tHit@5}}
\label{subsec:eval-adjusted}
Of the various systems whose performance we seek to compare, \textsf{QUEST}
and \our\ both produce ranked lists of answers that may contain ties. The
figures reported above are thus likely to over-estimate the accuracy of
these two systems. Table~\ref{tab:ties} shows an example explaining this
possibility. In the example, the answers returned by \textsf{QUEST}
contains 20 entities at rank 2, all of which are incorrect, but because a
correct answer is retrieved at rank 5 (before ties are broken), the values
of traditional \metric{MRR} and \metric{Hit@5} values are 0.2 and 1,
respectively.

In this section, therefore, we compare the performance of the best systems
using tie-aware versions of the above evaluation measures, viz.,
\metric{tMRR}, \metric{tP@1} and \metric{tHit@5}. The values of these
metrics for \textsf{DrQA}, \textsf{QUEST} and \our\ are shown in
Table~\ref{baseline_cq_w_t_new_metric_plus_ourbest}. Note that
\textsf{DrQA}, \textsf{ShortestPaths} and \textsf{BFS} return \emph{single}
answers at each rank; the values of the conventional and tie-aware
metrics would be the same for these methods. We have excluded the
relatively weak baselines --- \textsf{ShortestPaths} and \textsf{BFS} ---
from the table.

Once again, the results show that that \our\ consistently outperforms both
\textsf{DrQA} and \textsf{QUEST} in terms of \metric{tMRR}, \metric{tP@1}
as well as \metric{tHit@5} on both sets of questions. The table also
confirms that traditional metrics grossly over-report the accuracy of
\textsf{QUEST}. When evaluated using tie-aware measures, \textsf{DrQA}'s
performance is observed to be significantly better than \textsf{QUEST}'s.
In hindsight, this is unsurprising, since \textsf{QUEST} produces lists
that often contain many different entities at a particular rank. Naturally,
the improvement of \our\ over \textsf{QUEST} is statistically significant
for all three metrics. More importantly, the \metric{tMRR} and
\metric{tHit@5} figures for \our\ are, almost without exception, also
statistically significantly higher than the corresponding figures for
\textsf{DrQA}.

\begin{figure*}[t]
  \centering

   \subfloat[CQ-W - \metric{MRR}, \metric{tMRR}] {\label{fig:mrr_cq_w}\includegraphics[width=0.16\textwidth]{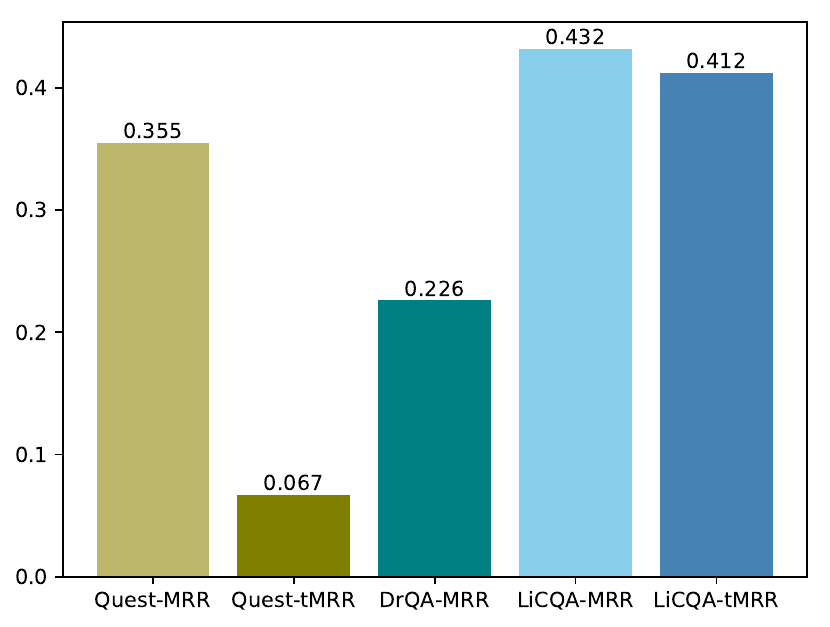}}
   \subfloat[CQ-W - \metric{P@1}, \metric{tP@1}] {\label{fig:p_1_cq_w}\includegraphics[width=0.16\textwidth]{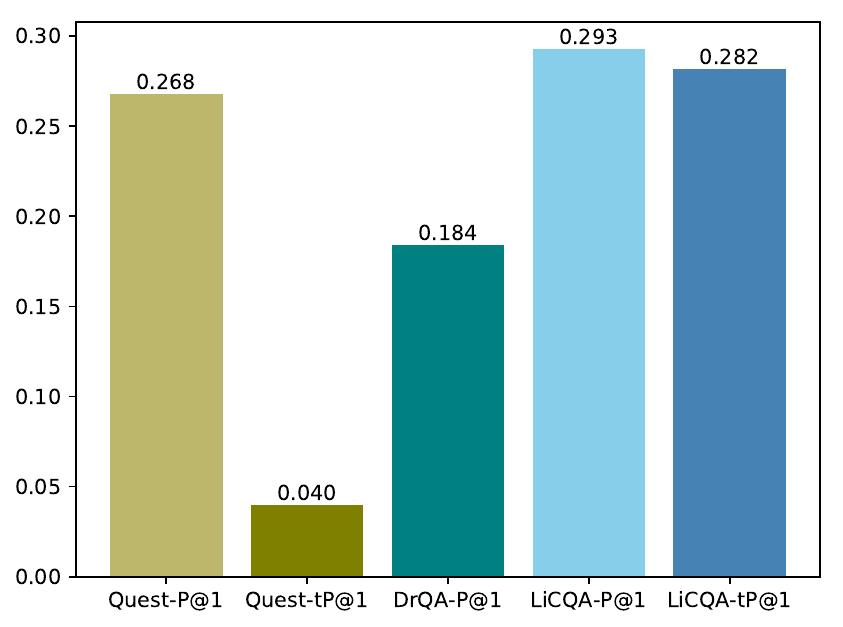}}
   \subfloat[CQ-W - \metric{Hit@5}, \metric{tHit@5}] {\label{fig:hit_cq_w}\includegraphics[width=0.16\textwidth]{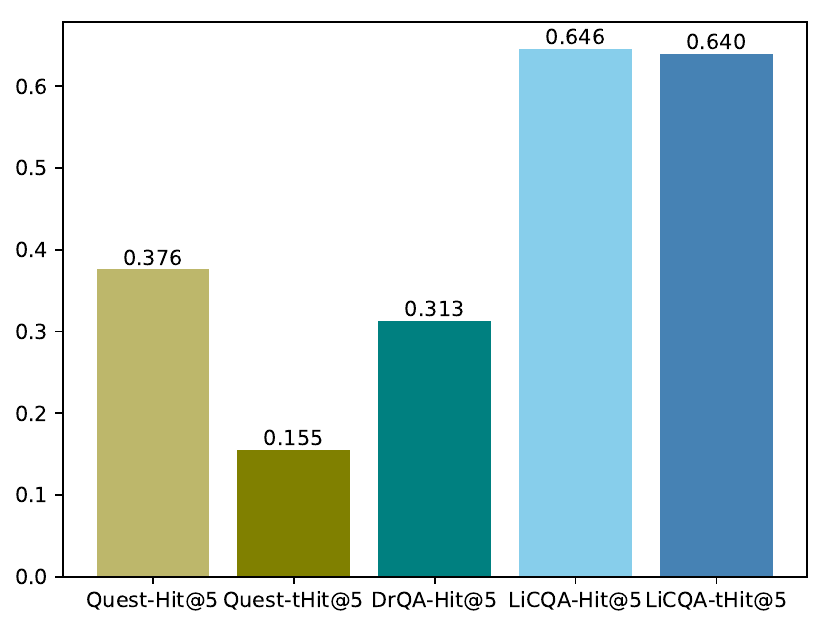}}
   \subfloat[CQ-T - \metric{MRR}, \metric{tMRR}] {\label{fig:mrr_cq_t}\includegraphics[width=0.16\textwidth]{figs/barplot_mrr_cq_w.pdf}}
  \subfloat[CQ-T - \metric{P@1}, \metric{tP@1}] {\label{fig:p_1_cq_t}\includegraphics[width=0.16\textwidth]{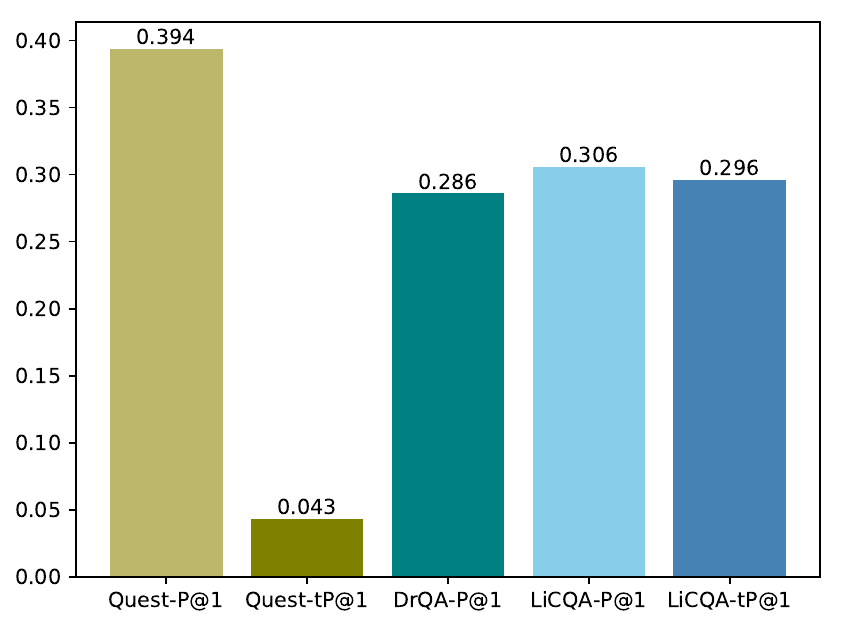}}
  \subfloat[CQ-T - \metric{Hit@5}, \metric{tHit@5}] {\label{fig:hit_cq_t}\includegraphics[width=0.16\textwidth]{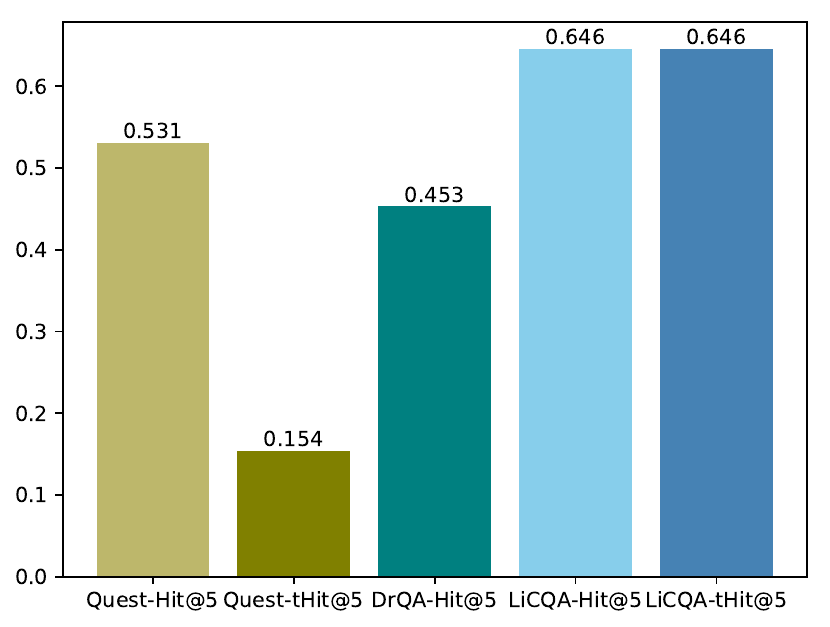}}
 
  \caption{Performance analysis of \our, \textsf{DrQA} and \textsf{QUEST} on CQ-W and CQ-T question set on the Top10 dataset. First and second barplots (shown in `khaki' and `olive' color) denote \textsf{QUEST} result with normal and tie-aware metrics respectively. 
  Similarly fourth and fifth barplots (shown in `skyblue' and `steelblue' color) denote \our~ result with normal and tie-aware metrics respectively. Middle bar with teal color presents result of \textsf{DrQA} (as it retrieves a single entity per rank position).
  \label{fig:bar-plot}}
\end{figure*}

Figure~\ref{fig:bar-plot} graphically compares the performance of \our,
\textsf{DrQA} and \textsf{QUEST} using both conventional and tie-aware
metrics on CQ-W and CQ-T for the Top10 collection. Each plot contains five
bars that correspond respectively to each of the classical and tie-aware
metrics for \textsf{QUEST} (in \emph{khaki} and \emph{olive}),
\textsf{DrQA} (\emph{teal}) and \our\ (in \emph{skyblue} and
\emph{steelblue}). As noted before, \textsf{DrQA} always retrieves a single
entity at any rank, so both metrics have the same value. From the figure,
it is clear that the skyblue bars are generally the tallest, except for P@1
on CQ-T, where \textsf{QUEST} performed the best. For the tie-aware
metrics, \our\ consistently achieves the best performance (steelblue bar)
in all cases.

\begin{table}[h]
\centering
\resizebox{0.999999\columnwidth}{!}
{
\begin{tabular}{c L c c c c}
\toprule
Question Set & Question &  Answer & \our\ rank & QUEST rank & DrQA rank  \\ [0.9ex] 
\hline 
\multirow{3}{*}{CQ-W} & What movie starred Bruce Willis and Haley Joel Osment?  & The Sixth Sense & \textbf{1} & $>$5 & 3 \\
\cmidrule{2-6}
& Who founded Apple and Pixar? & Steve Jobs & \textbf{1} & 2 & 5\\
\hline
\multirow{4}{*}{CQ-T} & Which 2018 studio album is performed by Playboi Carti and features Nicki Minaj?  & Die Lit  & \textbf{1} & $>$5 & 2 \\
\cmidrule{2-6}
& Which American professional golfer is married to Taylor Dowd Simpson and played for Wake Forest University? & Simpson  & \textbf{1} & $>$5 & $>$5 \\
\bottomrule
\end{tabular}
}
\caption{Some example questions for which \our\ has been able to retrieve the correct answer at rank 1 where as \textsf{QUEST} and \textsf{DrQA} have failed to retrieve the answers in the top rank.
}
\label{tab:our_best_examples}
\end{table}

\begin{figure*}[t]
  \centering
%
%
%
%
%
%
   \subfloat[Diff. in \metric{MRR}] {\label{fig:diff-w-mrr}\includegraphics[width=0.16\textwidth]{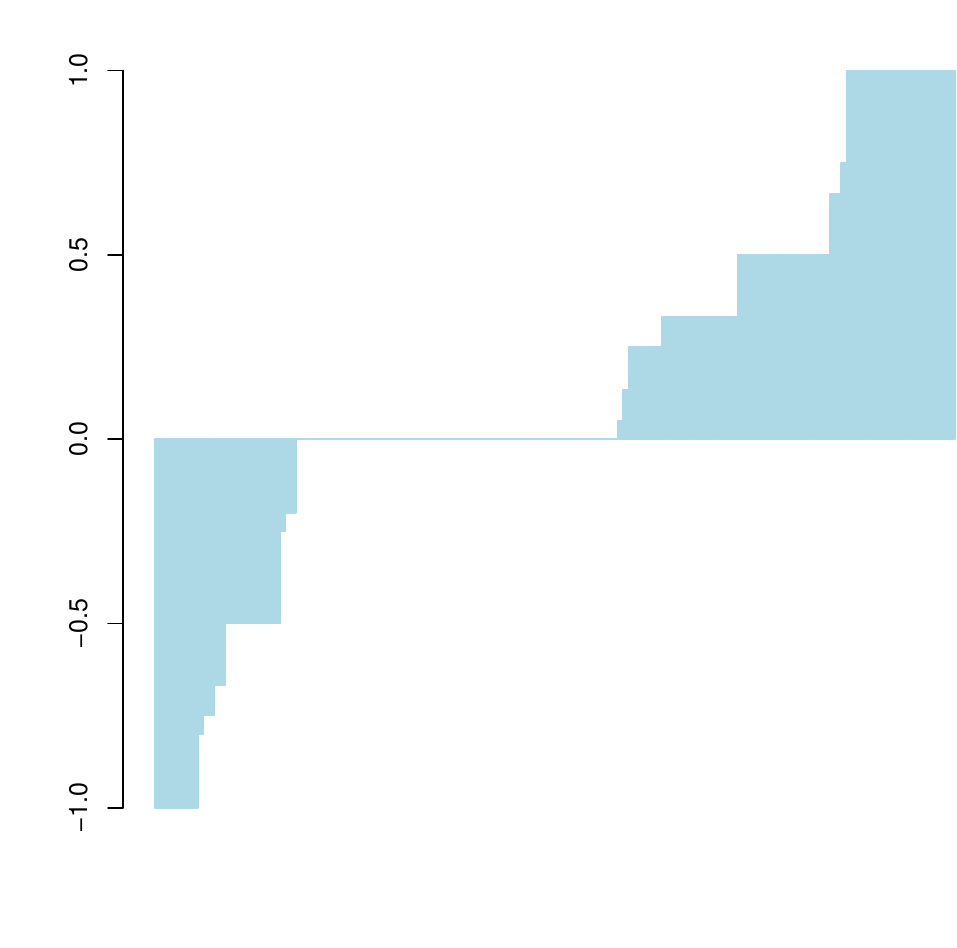}}
   \subfloat[Diff. in \metric{P@1}] {\label{fig:diff-w-p1}\includegraphics[width=0.16\textwidth]{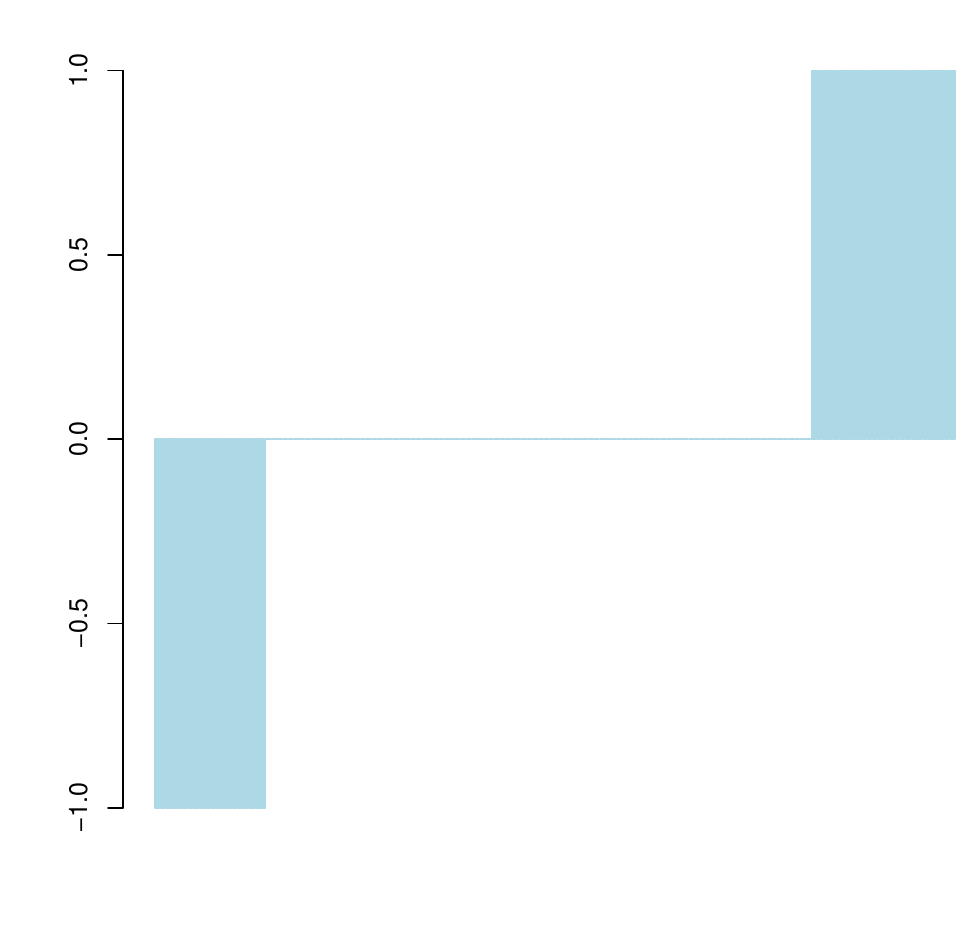}}
   \subfloat[Diff. in \metric{Hit@5}] {\label{fig:diff-w-h5}\includegraphics[width=0.16\textwidth]{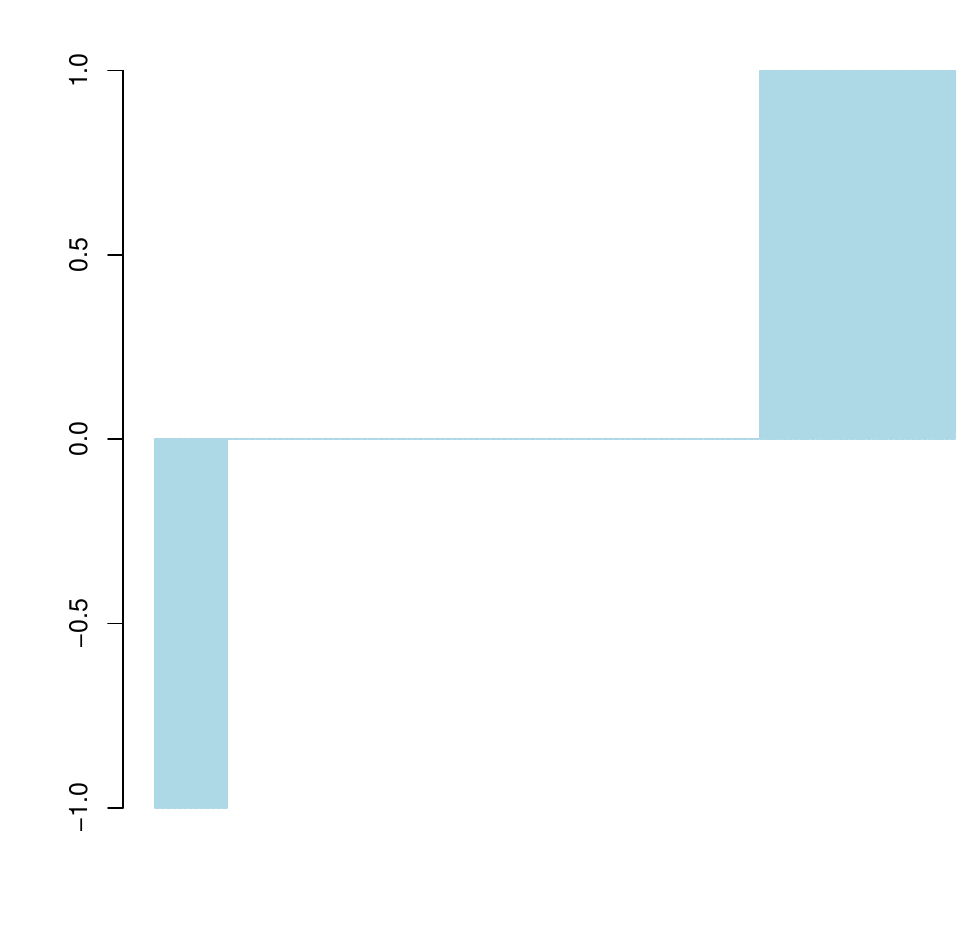}}
   \subfloat[Diff. in \metric{tMRR}] {\label{fig:diff-w-amrr}\includegraphics[width=0.16\textwidth]{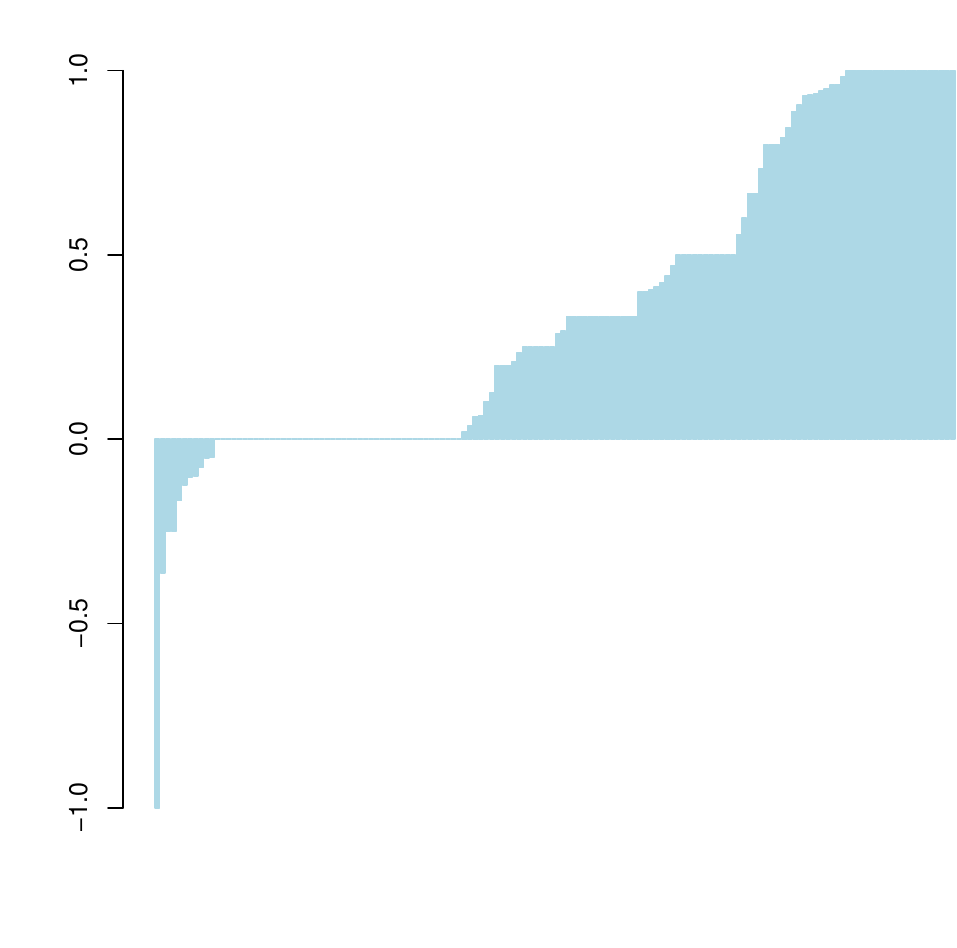}}
  \subfloat[Diff. in \metric{tP@1}] {\label{fig:diff-w-ap1}\includegraphics[width=0.16\textwidth]{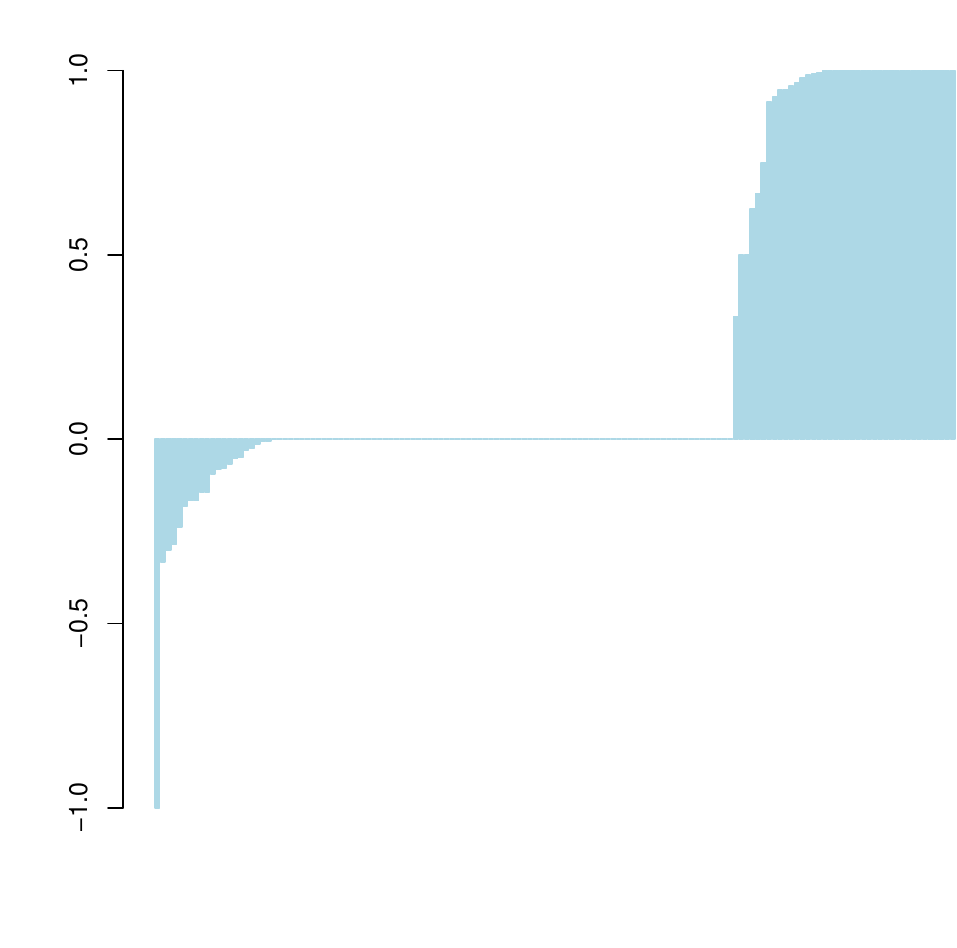}}
  \subfloat[Diff. in \metric{tHit@5}] {\label{fig:diff-w-ah5}\includegraphics[width=0.16\textwidth]{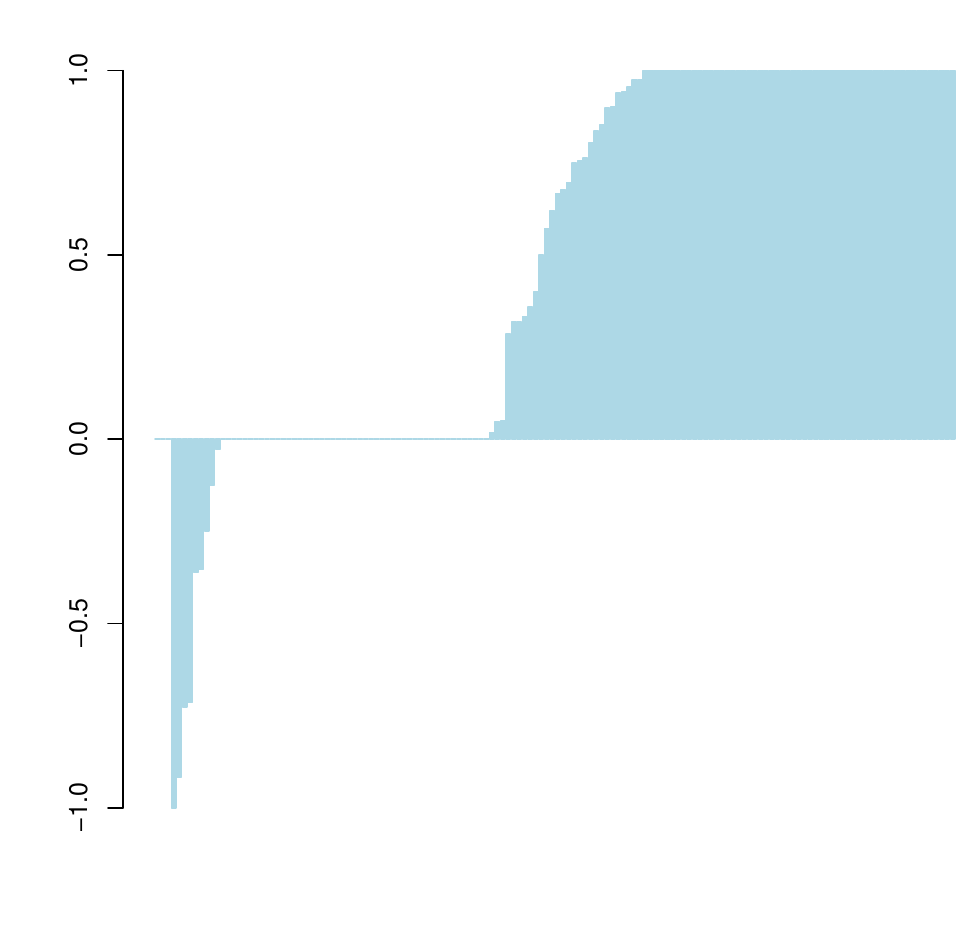}}
 
  \caption{A per-query performance analysis between \our~and \textsf{QUEST} on CQ-W question set for Top10 dataset. 
  Bars above x-axis correspond to questions for which \our~has achieved better performance than QUEST. \label{fig:per-query-cqw}}
\end{figure*}

\begin{figure*}[t]
  \centering
%
%
%
%
%
  \subfloat[Diff. in \metric{MRR}] {\label{fig:diff-t-mrr}\includegraphics[width=0.16\textwidth]{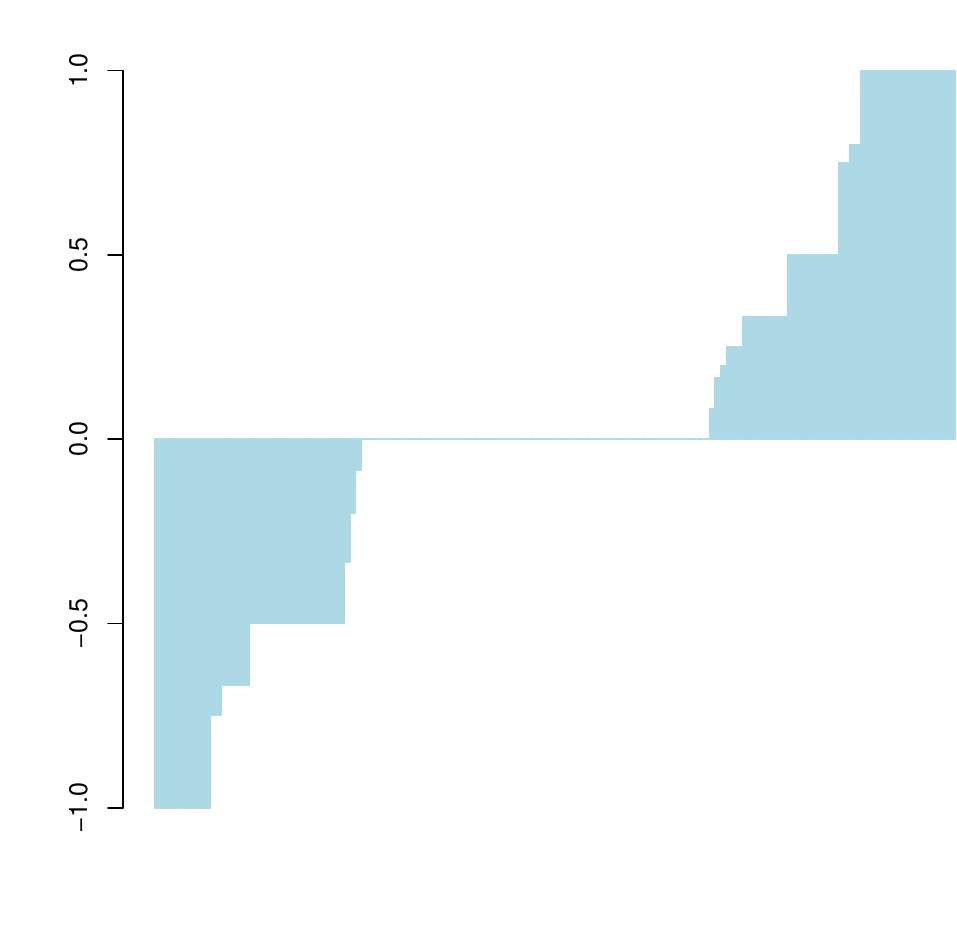}}
  \subfloat[Diff. in \metric{P@1}] {\label{fig:diff-t-p1}\includegraphics[width=0.16\textwidth]{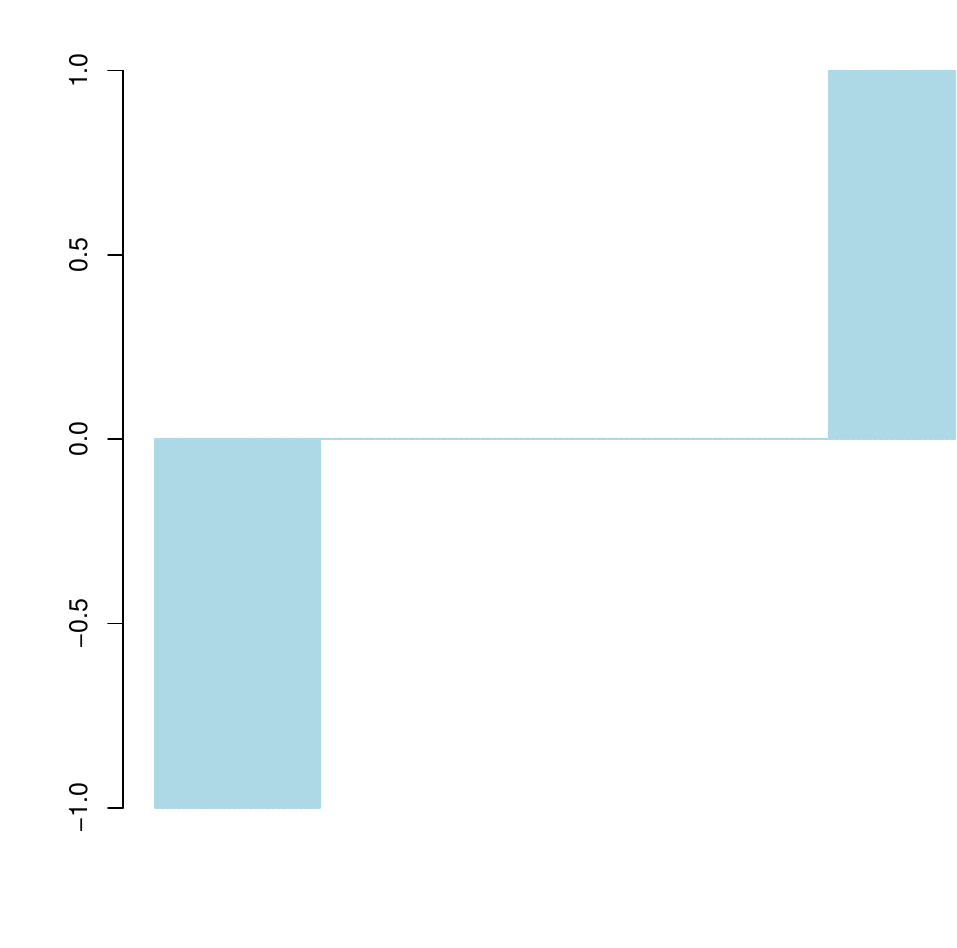}}
  \subfloat[Diff. in \metric{Hit@5}] {\label{fig:diff-t-h5}\includegraphics[width=0.16\textwidth]{figs/qq-diffs-our_vs_quest_hit_t_new-crop.pdf}}
  \subfloat[Diff. in \metric{tMRR}] {\label{fig:diff-t-amrr}\includegraphics[width=0.16\textwidth]{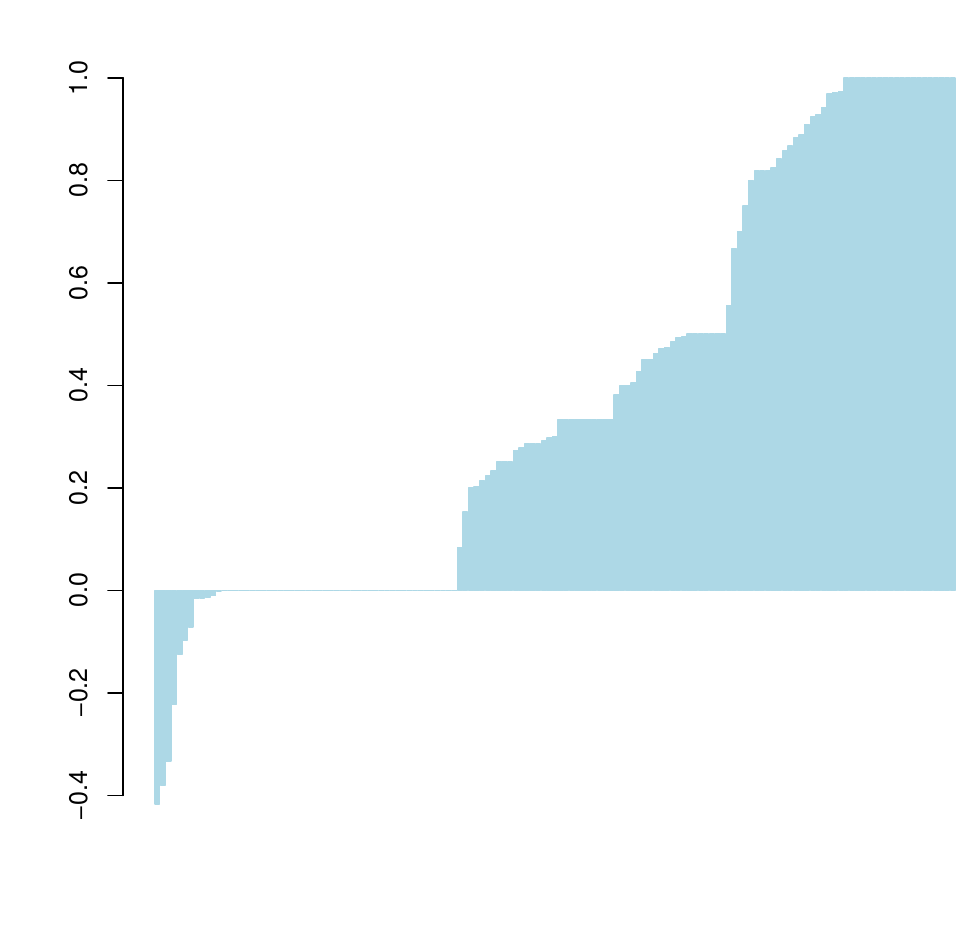}}
  \subfloat[Diff. in \metric{tP@1}] {\label{fig:diff-t-ap1}\includegraphics[width=0.16\textwidth]{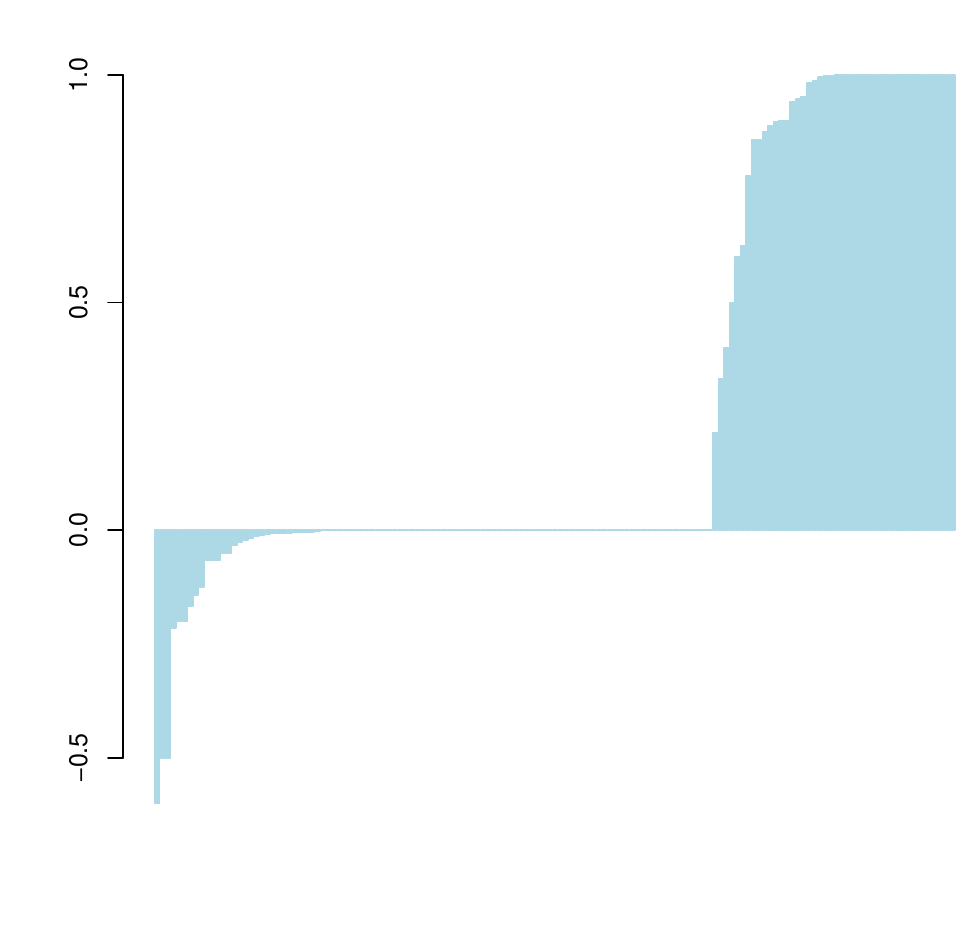}} 
  \subfloat[Diff. in \metric{tHit@5}] {\label{fig:diff-t-ah5}\includegraphics[width=0.16\textwidth]{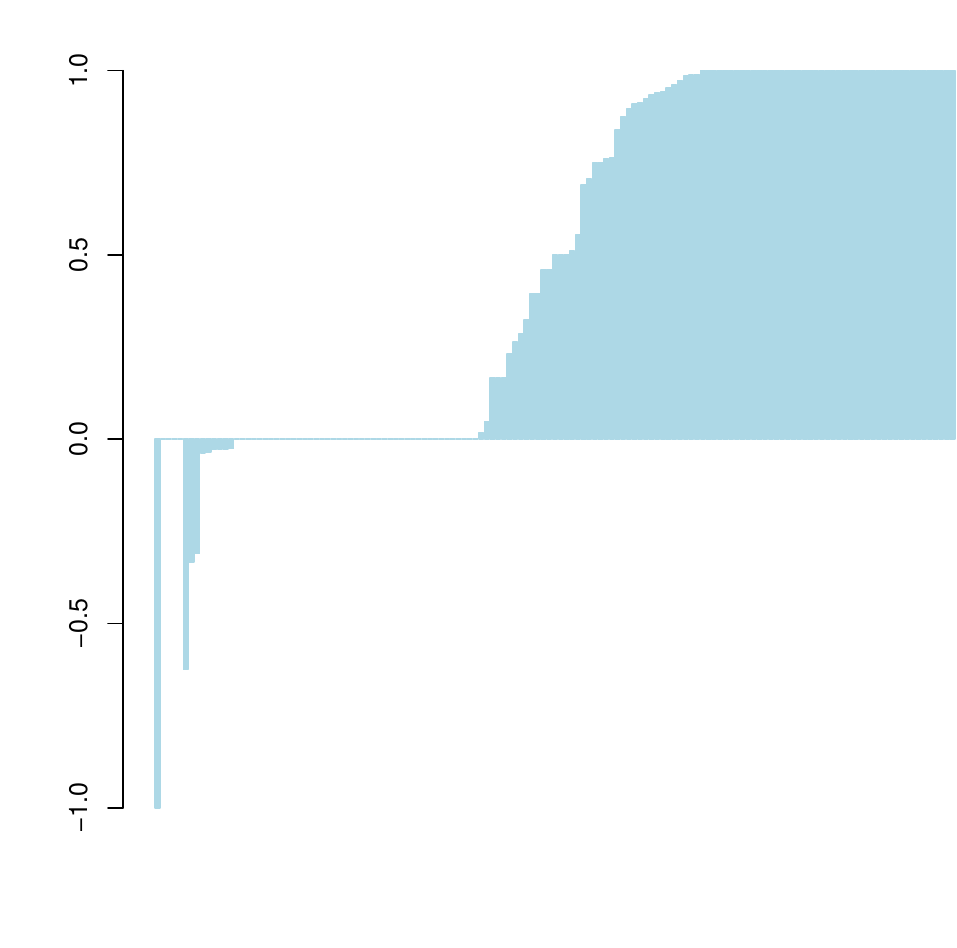}}
  \caption{A per-query performance analysis between \our~and \textsf{QUEST} on CQ-T question set for Top10 dataset. 
  Bars above x-axis correspond to questions for which \our~has achieved better performance than QUEST.\label{fig:per-query-cqt}}
\end{figure*}

\subsection{Qualitative comparison}
Next, we take a closer look at the relative performance of \textsf{QUEST}
and \our. Figures~\ref{fig:per-query-cqw} and~\ref{fig:per-query-cqt} show
the per-query performance difference between the two systems
on the Top10 dataset. In these figures, a bar above the x-axis indicates
that \our\ performed better than \textsf{QUEST} for the corresponding
question. Figures~\ref{fig:diff-w-mrr}--\ref{fig:diff-w-h5} and
Figures~\ref{fig:diff-t-mrr}-\ref{fig:diff-t-h5}) show that \our\ performs
better for a majority of the questions on both sets. Note that, for a
single question, the values of \metric{P@1} and \metric{Hit@5} are binary
(0 or 1), and the difference in these values can be $1$, $0$ or $-1$.

In contrast, the tie-aware measures have non-binary values. The per-query
difference with respect to these metrics are shown in
Figure~\ref{fig:diff-w-amrr}-\ref{fig:diff-w-ah5} and
Figure~\ref{fig:diff-t-amrr}-\ref{fig:diff-t-ah5}, for CQ-W and CQ-T,
respectively. Once again, it is clear from these figures that \our\
outperforms \textsf{QUEST} for most questions.

Some example questions for which \our~\ was able to retrieve the correct
answer at top rank, but \textsf{QUEST} and \textsf{DrQA} failed, are
presented in Table~\ref{tab:our_best_examples}. 



\subsection{Latency analysis}
Finally, we compare \our\ with \textsf{DrQA} and \textsf{QUEST} in terms of
the average time taken to produce the list of answers for a given set of
questions. Each system is made to run in a loop for five iterations, and
computes the answers for all questions in CQ-W and CQ-T in every iteration.
The average of the time taken over these five iterations is reported in
Table~\ref{table:time_comparison}.

All our experiments were conducted on an Intel(R) Core(TM) i9-7900X
CPU@3.30GHz and 24 GB Nvidia Titan RTX GPU machine with 132 GB RAM. Among
the methods, \textsf{QUEST} takes the highest time in producing the
results, spending a significant time in creating the quasi knowledge
graph with the retrieved documents. The latency of \textsf{DrQA} is
also close to \textsf{QUEST}. In contrast, the execution time
of \our\ is significantly less than both the baselines. More specifically,
we observe an $8\times$ speedup for \our\ compared to the baselines in the
total running time taken to process the two question sets for all the
document sets (Top10 and Strata1--5).

\begin{table}[h!]
\resizebox{0.99\columnwidth}{!}
{
\begin{tabular}{c c c c c c c c} 
\hline
    System &    Query set    & Top10  & Strata-1 & Strata-2 & Strata-3 & Strata-4 & Strata-5 \\
\hline
\textsf{\our}   &  \multirow{3}{*}{CQ-W}    &   \textbf{13}  & \textbf{15}  &  \textbf{17} & \textbf{14} & \textbf{16.5} & \textbf{14.3} \\
\textsf{QUEST}  &                           &   144 &  148 & 151 & 156 & 149 & 162 \\
\textsf{DrQA}   &                           &   120 &  131 & 119 & 124.6 & 128 & 129\\ [0.5ex]

\hline    
speedup  &                                  &   9.23$\times$    &  8.73$\times$      &  7$\times$  &  8.9$\times$     & 7.75$\times$  & 9.02$\times$ \\[0.5ex]
\hline

\textsf{\our}   &  \multirow{3}{*}{CQ-T}    &   \textbf{14.5}  & \textbf{16} & \textbf{18} & \textbf{15.8} & \textbf{18} & \textbf{17.6} \\
\textsf{QUEST}  &                           &   155   &  159 & 162 & 178 & 189.5 & 195.4  \\
\textsf{DrQA}   &                           &   130   &  135 & 138.6 & 132.8 & 142 & 137  \\ [0.5ex]
\hline
speedup  &                                  &   8.96$\times$ &   8.43$\times$ & 7.7$\times$  & 8.40$\times$ & 7.88$\times$ & 7.78$\times$\\[0.5ex]
\hline

\end{tabular}
}
\caption{Per query latency (in second) analysis of different QA models. Least latency highlighted in bold. Speedup row indicates the reduction in execution time than the baseline with the minimum latency (DrQA).}
\label{table:time_comparison}
\end{table}

\section{Conclusion and Future Work} \label{con_future_work}
In this paper, we present~\our, a low-resource, unsupervised, complex question answering system that uses 
various forms of corpus evidence.  
To evaluate \our, experiments were conducted on diverse question sets taken from public forums like WikiAnswers and Google Trends.
\our\ was compared with state-of-the-art systems \textsf{DrQA} and \textsf{QUEST} using \metric{MRR}, \metric{P@1}, \metric{Hit@5}, and their tie-aware variants. 
\our\ achieved significant improvements over all baselines in a majority of
cases, while simultaneously reducing computational cost (as measured by the
average time taken to answer a question).

Recently developed neural models for representing complex concepts via low dimensional vectors~\cite{ir/ShalabyZJ19,wikipedia2vec2020} have been successfully applied to different text processing and information extraction tasks~\cite{application_word_2_sense}.
In future, we plan to explore such embedding models to further improve our
answer extraction module.

\tiny{
\bibliographystyle{ACM-Reference-Format}
\bibliography{refs}

@inproceedings{trec_ques_classification_data,
        author = "Li, Xin and 
        Roth, Dan",
        title = {Learning Question Classifiers},
        year = {2002},
        booktitle = {Proc. of 19th {COLING}},
        pages = {1–7},
numpages = {7},
series = {COLING ’02}
}

@inproceedings{class_what_ques,
    title = {Classifying What-Type Questions by Head Noun Tagging},
    author = "Li, Fangtao and 
    Zhang, Xian and 
    Yuan, Jinhui and 
    Zhu, Xiaoyan",
    booktitle = "Proc. of 22nd Coling ",
    year = 2008,
    pages = "481--488",
}

@InProceedings{infersent,
  author    = {Conneau, Alexis  and  Kiela, Douwe  and  Schwenk, Holger  and  Barrault, Lo\"{i}c  and  Bordes, Antoine},
  title     = {Supervised Learning of Universal Sentence Representations from Natural Language Inference Data},
  booktitle = {Proc. of 2017 {EMNLP}},
  month     = {September},
  year      = {2017},
  pages     = {670--680}
}

@inproceedings{quest,
author = {Lu, Xiaolu and Pramanik, Soumajit and Saha Roy, Rishiraj and Abujabal, Abdalghani and Wang, Yafang and Weikum, Gerhard},
title = {Answering Complex Questions by Joining Multi-Document Evidence with Quasi Knowledge Graphs},
year = {2019},
isbn = {9781450361729},


booktitle = {Proc. of 42nd {SIGIR}},
pages = {105–114},
numpages = {10},
series = {SIGIR’19}
}

@inproceedings{sentence-bert,
    title = "Sentence-BERT: Sentence Embeddings using Siamese BERT-Networks",
    author = "Reimers, Nils and Gurevych, Iryna",
    booktitle = "Proc. of 2019 {EMNLP}",
    month = "11",
    year = "2019",
    pages = "3982--3992",

}

@inproceedings{chen2017reading,
  title={Reading {Wikipedia} to Answer Open-Domain Questions},
  author={Chen, Danqi and Fisch, Adam and Weston, Jason and Bordes, Antoine},
  booktitle={Proceedings of 55th {ACL} 2017},
  year={2017},
  pages = "1870--1879",

}

@article{prager,
author = {Prager, John},
title = {Open-Domain Question Answering},
year = {2006},
issue_date = {January 2006},
volume = {1},
number = {2},
issn = {1554-0669},
journal = {Found. Trends Inf. Retr.},
month = jan,
pages = {91–231},
numpages = {141}
}

@inproceedings{unsupervised-qa,
    title = "Unsupervised Question Answering by Cloze Translation",
    author = "Patrick Lewis  and
      Ludovic Denoyer  and
      Sebastian Riedel",
    booktitle = "Proc. of 57th {ACL}",
    month = jul,
    year = "2019",
    pages = "4896--4910"
}

@inproceedings{squad,
    title = "{SQ}u{AD}: 100,000+ Questions for Machine Comprehension of Text",
    author = "Rajpurkar, Pranav  and
      Zhang, Jian  and
      Lopyrev, Konstantin  and
      Liang, Percy",
    booktitle = "Proc. of 2016 {EMNLP}",
    month = nov,
    year = "2016",
    
    pages = "2383--2392",
}

@inproceedings{wikimovies,
    title = "Key-Value Memory Networks for Directly Reading Documents",
    author = "Miller, Alexander  and
      Fisch, Adam  and
      Dodge, Jesse  and
      Karimi, Amir-Hossein  and
      Bordes, Antoine  and
      Weston, Jason",
    booktitle = "Proc. 2016 {EMNLP}",
    month = nov,
    year = "2016",
    
    pages = "1400--1409",
}

@article{wikipedia2vec2020,
  title={Wikipedia2Vec: An Efficient Toolkit for Learning and Visualizing the Embeddings of Words and Entities from Wikipedia},
  author={Yamada, Ikuya and Asai, Akari and Sakuma, Jin and Shindo, Hiroyuki and Takeda, Hideaki and Takefuji, Yoshiyasu and Matsumoto, Yuji},
  journal={arXiv preprint 1812.06280v3},
  year={2020}
}

@inproceedings{flair,
  title={Pooled Contextualized Embeddings for Named Entity Recognition},
  author={Akbik, Alan and Bergmann, Tanja and Vollgraf, Roland},
  booktitle = {{NAACL} 2019},
  pages     = {724 - 728},
  year      = {2019}
}

@inproceedings{cer-etal-2018-universal,
    title = "Universal Sentence Encoder for {E}nglish",
    author = "Cer, Daniel  and
      Yang, Yinfei  and
      Kong, Sheng-yi  and
      Hua, Nan  and
      Limtiaco, Nicole  and
      St. John, Rhomni  and
      Constant, Noah  and
      Guajardo-Cespedes, Mario  and
      Yuan, Steve  and
      Tar, Chris  and
      Strope, Brian  and
      Kurzweil, Ray",
    booktitle = "Proc. 2018 {EMNLP}: System Demonstrations",
    month = nov,
    year = "2018",
    
    pages = "169--174"
}

@inproceedings{bowman-etal-2015-large,
    title = "A large annotated corpus for learning natural language inference",
    author = "Bowman, Samuel R.  and
      Angeli, Gabor  and
      Potts, Christopher  and
      Manning, Christopher D.",
    booktitle = "Proc. 2015 {EMNLP}",
    month = sep,
    year = "2015",
    
    pages = "632--642",
}

@article{qa_type_survey,
  year = {2010},
  month = nov,
  volume = {35},
  number = {2},
  pages = {137--154},
  author = {Jo{\~{a}}o Silva and Lu{\'{\i}}sa Coheur and Ana Cristina Mendes and Andreas Wichert},
  title = {From symbolic to sub-symbolic information in question classification},
  journal = {Artificial Intelligence Review}
}

@inproceedings{bert,
    title = "{BERT}: Pre-training of Deep Bidirectional Transformers for Language Understanding",
    author = "Devlin, Jacob  and
      Chang, Ming-Wei  and
      Lee, Kenton  and
      Toutanova, Kristina",
    booktitle = "Proc. of {NAACL}",
    month = jun,
    year = "2019",
    
    pages = "4171--4186",
}

@article{roberta,
  author    = {Yinhan Liu and
               Myle Ott and
               Naman Goyal and
               Jingfei Du and
               Mandar Joshi and
               Danqi Chen and
               Omer Levy and
               Mike Lewis and
               Luke Zettlemoyer and
               Veselin Stoyanov},
  title     = {RoBERTa: {A} Robustly Optimized {BERT} Pretraining Approach},
  journal   = {CoRR},
  volume    = {abs/1907.11692},
  year      = {2019},
  archivePrefix = {arXiv},
  eprint    = {1907.11692},
  bibsource = {dblp computer science bibliography, https://dblp.org}
}

@inproceedings{hotpotqa,
  title={{HotpotQA}: A Dataset for Diverse, Explainable Multi-hop Question Answering},
  author={Yang, Zhilin and Qi, Peng and Zhang, Saizheng and Bengio, Yoshua and Cohen, William W. and Salakhutdinov, Ruslan and Manning, Christopher D.},
  booktitle={{EMNLP}},
  year={2018},
  pages = "2369--2380",

}

@inproceedings{docqa,
    title = "Simple and Effective Multi-Paragraph Reading Comprehension",
    author = "Clark, Christopher  and
      Gardner, Matt",
    booktitle = "Proc. of 56th {ACL}",
    month = jul,
    year = "2018",
    
    pages = "845--855",
}

@inproceedings{factoid_kg_qa,
author = {Zhao, Chen and Xiong, Chenyan and Qian, Xin and Boyd-Graber, Jordan},
title = {Complex Factoid Question Answering with a Free-Text Knowledge Graph},
year = {2020},
isbn = {9781450370233},


booktitle = {Proc. of Web Conference 2020},
pages = {1205–1216},
numpages = {12},
series = {WWW ’20}
}

@INPROCEEDINGS{Voorhees99thetrec-8,
    author = {Ellen M. Voorhees},
    title = {The TREC-8 Question Answering Track Report},
    booktitle = {In Proc. of TREC-8},
    year = {1999},
    pages = {77--82}
}

@INPROCEEDINGS {bfs_from_quest,
author = {F. M. Suchanek and G. Weikum and M. Ramanath and G. Kasneci and M. Sozio},
booktitle = {2013 IEEE 29th {ICDE}},
title = {STAR: Steiner-Tree Approximation in Relationship Graphs},
year = {2009},
volume = {},
issn = {1084-4627},
pages = {868-879},
month = {apr}
}

@inproceedings{qa_old_tois,
    title = "Performance Issues and Error Analysis in an Open-Domain Question Answering System",
    author = "Moldovan, Dan  and
      Pasca, Marius  and
      Harabagiu, Sanda  and
      Surdeanu, Mihai",
    booktitle = "Proc. of 40th {ACL}",
    month = jul,
    year = "2002",
    
    pages = "33--40",
}

@inproceedings{singhal:trec8,
  author    = {Amit Singhal and
               Steven P. Abney and
               Michiel Bacchiani and
               Michael Collins and
               Donald Hindle and
               Fernando C. N. Pereira},
  title     = {AT{\&}T at {TREC-8}},
  booktitle = {Proc. of Eighth {TREC} 1999},
  volume    = {500-246},
  year      = {1999},
}

@inproceedings{lasso:trec8,
  author    = {Dan Moldovan and
               Sanda Harabagiu and
               Marius Pasca and
               Rada Mihalcea and
               Richard Goodrum and
               Roxana Girju and
               Vasile Rus},
  title     = {{LASSO:} {A} Tool for Surfing the Answer Net},
  booktitle = {Proc. of Eighth {TREC} 1999},
  volume    = {500-246},
  year      = {1999}
}

@inproceedings{srihari:ie,
  author    = {Rohini K. Srihari and
               Wei Li},
  title     = {A Question Answering System Supported by Information Extraction},
  booktitle = {6th {ANLP}, Seattle,
               Washington, USA},
  pages     = {166--172},
  year      = {2000},
}

@inproceedings{dong:cnn,
  author    = {Li Dong and
               Furu Wei and
               Ming Zhou and
               Ke Xu},
  title     = {Question Answering over Freebase with Multi-Column Convolutional Neural Networks},
  booktitle = {Proc. of 53rd {ACL}},
  pages     = {260--269},
  year      = {2015},
}

@article{tan:rnn,
  author    = {Chuanqi Tan and
               Furu Wei and
               Qingyu Zhou and
               Nan Yang and
               Bowen Du and
               Weifeng Lv and
               Ming Zhou},
  title     = {Context-Aware Answer Sentence Selection With Hierarchical Gated Recurrent
               Neural Networks},
  journal   = {{IEEE} {ACM} Trans. Audio Speech Lang. Process.},
  volume    = {26},
  number    = {3},
  pages     = {540--549},
  year      = {2018},
}

@inproceedings{lrec/BuscaldiR06,
  author    = {Davide Buscaldi and
               Paolo Rosso},
  title     = {Mining Knowledge from Wikipedia for the Question Answering task},
  booktitle = {Proc. of Fifth {LREC}},
  pages     = {727--730},
  year      = {2006},
}

@inproceedings{trec/AhnJMMRS04,
  author    = {David Ahn and
               Valentin Jijkoun and
               Gilad Mishne and
               Karin M{\"{u}}ller and
               Maarten de Rijke and
               Stefan Schlobach},
  title     = {Using Wikipedia at the {TREC} {QA} Track},
  booktitle = {Proc. of Thirteenth {TREC}},
  series    = {{NIST} Special Publication},
  volume    = {500-261},
  year      = {2004}
}

@inproceedings{UsbeckNHKRN17,
  author    = {Ricardo Usbeck and
               Axel{-}Cyrille Ngonga Ngomo and
               Bastian Haarmann and
               Anastasia Krithara and
               Michael R{\"{o}}der and
               Giulio Napolitano},
  title     = {7th Open Challenge on Question Answering over Linked Data {(QALD-7)}},
  pages     = {59--69},
  year      = {2017},
}

@inproceedings{berant-etal-2013-semantic,
    title = "Semantic Parsing on {F}reebase from Question-Answer Pairs",
    author = "Berant, Jonathan  and
      Chou, Andrew  and
      Frostig, Roy  and
      Liang, Percy",
    booktitle = "Proc. of 2013 {EMNLP}",
    month = oct,
    year = "2013",
    pages = "1533--1544",
}

@inproceedings{naacl/TalmorB18,
  author    = {Alon Talmor and
               Jonathan Berant},
  title     = {The Web as a Knowledge-Base for Answering Complex Questions},
  booktitle = {Proc. of {NAACL-HLT}},
  pages     = {641--651},
  year      = {2018},
}

@inproceedings{acl/KhotSC17,
  author    = {Tushar Khot and
               Ashish Sabharwal and
               Peter Clark},
  title     = {Answering Complex Questions Using Open Information Extraction},
  booktitle = {Proc. of 55th {ACL} 2017},
  pages     = {311--316},
  year      = {2017},
  bibsource = {dblp computer science bibliography, https://dblp.org}
}

@inproceedings{acl/FaderZE13,
  author    = {Anthony Fader and
               Luke S. Zettlemoyer and
               Oren Etzioni},
  title     = {Paraphrase-Driven Learning for Open Question Answering},
  booktitle = {Proc. of 51st {ACL}},
  pages     = {1608--1618},
  year      = {2013},
}

@inproceedings{kdd/FaderZE14,
  author    = {Anthony Fader and
               Luke Zettlemoyer and
               Oren Etzioni},
  title     = {Open question answering over curated and extracted knowledge bases},
  booktitle = {The 20th {ACM} {SIGKDD}},
  pages     = {1156--1165},
  year      = {2014},
}

@inproceedings{sigir/SavenkovA16,
  author    = {Denis Savenkov and
               Eugene Agichtein},
  title     = {When a Knowledge Base Is Not Enough: Question Answering over Knowledge
               Bases with External Text Data},
  booktitle = {Proc. of 39th{SIGIR}},
  pages     = {235--244},
  year      = {2016},
  bibsource = {dblp computer science bibliography, https://dblp.org}
}

@inproceedings{acl/CaiY13,
  author    = {Qingqing Cai and
               Alexander Yates},
  title     = {Large-scale Semantic Parsing via Schema Matching and Lexicon Extension},
  booktitle = {Proc. of 51st {ACL}},
  pages     = {423--433},
  year      = {2013}
}

@inproceedings{www/AbujabalYRW17,
  author    = {Abdalghani Abujabal and
               Mohamed Yahya and
               Mirek Riedewald and
               Gerhard Weikum},
  title     = {Automated Template Generation for Question Answering over Knowledge
               Graphs},
  booktitle = {Proc. of 26th {WWW}},
  pages     = {1191--1200},
  year      = {2017},
}

@inproceedings{cikm/BastH15,
  author    = {Hannah Bast and
               Elmar Haussmann},
  title     = {More Accurate Question Answering on Freebase},
  booktitle = {Proc. of 24th {ACM} {CIKM}},
  pages     = {1431--1440},
  year      = {2015},
}

@inproceedings{convqa_mpi,
author = {Christmann, Philipp and Saha Roy, Rishiraj and Abujabal, Abdalghani and Singh, Jyotsna and Weikum, Gerhard},
title = {Look before You Hop: Conversational Question Answering over Knowledge Graphs Using Judicious Context Expansion},
year = {2019},
booktitle = {Proc of 28th {ACM CIKM}},
pages = {729–738},
numpages = {10},
series = {CIKM ’19}
}

@article{ir/ShalabyZJ19,
  author    = {Walid Shalaby and
               Wlodek Zadrozny and
               Hongxia Jin},
  title     = {Beyond word embeddings: learning entity and concept representations
               from large scale knowledge bases},
  journal   = {Inf. Retr. J.},
  volume    = {22},
  number    = {6},
  pages     = {525--542},
  year      = {2019},
}

@inproceedings{application_word_2_sense,
  author    = {Irina Gruzdo and
               Iryna Kyrychenko and
               Glib Tereshchenko and
               Olga Cherednichenko},
  title     = {Appl{\i}cat{\i}on of Paragraphs Vectors Model for Semant{\i}c Text
               Analys{\i}s},
  booktitle = {Proc. of 4th {COLINS}},
  series    = {{CEUR} Workshop Proceedings},
  volume    = {2604},
  pages     = {283--293},
  year      = {2020},
}

@inproceedings{performance_open_domain_qa,
    title = "Performance Issues and Error Analysis in an Open-Domain Question Answering System",
    author = "Moldovan, Dan  and
      Pasca, Marius  and
      Harabagiu, Sanda  and
      Surdeanu, Mihai",
    booktitle = "Proceedings of 40th {ACL}",
    year = "2002",
    pages = "33--40",
}

@inproceedings{ques_class_log_linear,
author = {Blunsom, Phil and Kocik, Krystle and Curran, James R.},
title = {Question Classification with Log-Linear Models},
year = {2006},
booktitle = {Proceedings of 29th {SIGIR}},
pages = {615–616},
numpages = {2},
}

@inproceedings{high_acc_ques_class,
    title = "High Accuracy Rule-based Question Classification using Question Syntax and Semantics",
    author = "Tayyar Madabushi, Harish  and
      Lee, Mark",
    booktitle = "Proc. of the 26th {COLING} 2016",
    year = "2016",
    pages = "1220--1230",
}

@article{ques_class_grammar,
title = {Question categorization and classification using grammar based approach},
journal = {Information Processing \& Management},
volume = {54},
number = {6},
pages = {1228-1243},
year = {2018},
issn = {0306-4573},
author = {Alaa Mohasseb and Mohamed Bader-El-Den and Mihaela Cocea},

}

@inproceedings{najork_ties_paper,
author = {McSherry, Frank and Najork, Marc},
title = {Computing Information Retrieval Performance Measures Efficiently in the Presence of Tied Scores},
year = {2008},
booktitle = {Proc. of 30th ECIR},
pages = {414–421},
numpages = {8},
series = {ECIR'08}
}

@inproceedings{bhutani-etal-2020-answering,
    title = "Answering Complex Questions by Combining Information from Curated and Extracted Knowledge Bases",
    author = "Bhutani, Nikita  and
      Zheng, Xinyi  and
      Qian, Kun  and
      Li, Yunyao  and
      Jagadish, H.",
    booktitle = "Proc. of First Workshop on Natural Language Interfaces",
    month = jul,
    year = "2020",
    pages = "1--10"
}

@article{google_nqa,
title	= {Natural Questions: a Benchmark for Question Answering Research},
author	= {Tom Kwiatkowski and Jennimaria Palomaki and Olivia Redfield and Michael Collins and Ankur Parikh and Chris Alberti and Danielle Epstein and Illia Polosukhin and Matthew Kelcey and Jacob Devlin and Kenton Lee and Kristina N. Toutanova and Llion Jones and Ming-Wei Chang and Andrew Dai and Jakob Uszkoreit and Quoc Le and Slav Petrov},
year	= {2019},
journal	= {TACL}
}

@article{ontonote5,
  title={Ontonotes release 5.0},
  author={Ralph Weischedel and Martha Palmer and Mitchell Marcus and Eduard Hovy and Sameer Pradhan and Lance Ramshaw and Nianwen Xue and Ann Taylor and Jeff Kaufman and Michelle Franchini and Mohammed El-Bachouti and Robert Belvin and Ann Houston},
  journal={LDC2011T03, Philadelphia, Penn.: Linguistic Data Consortium},
  year={2013}
}

@article{lin-reusable-test-collections,
author = {Lin, Jimmy and Katz, Boris},
title = {Building a Reusable Test Collection for Question Answering},
year = {2006},
issue_date = {May 2006},
publisher = {John Wiley & Sons, Inc.},
address = {USA},
volume = {57},
number = {7},
issn = {1532-2882},
journal = {J. Am. Soc. Inf. Sci. Technol.},
month = may,
pages = {851–861},
numpages = {11}
}

@article{quant-eval-jaist,
  author    = {Nina Wacholder and
               Diane Kelly and
               Paul B. Kantor and
               Robert Rittman and
               Ying Sun and
               Bing Bai and
               Sharon G. Small and
               Boris Yamrom and
               Tomek Strzalkowski},
  title     = {A model for quantitative evaluation of an end-to-end question-answering
               system},
  journal   = {J. Assoc. Inf. Sci. Technol.},
  volume    = {58},
  number    = {8},
  pages     = {1082--1099},
  year      = {2007},
  url       = {https://doi.org/10.1002/asi.20560},
  doi       = {10.1002/asi.20560},
  bibsource = {dblp computer science bibliography, https://dblp.org}
}

@article{sawant-garg-chak-rama,
author = {Sawant, Uma and Garg, Saurabh and Chakrabarti, Soumen and Ramakrishnan, Ganesh},
title = {Neural Architecture for Question Answering Using a Knowledge Graph and Web Corpus},
year = {2019},
issue_date = {August 2019},
publisher = {Kluwer Academic Publishers},
address = {USA},
volume = {22},
number = {3–4},
issn = {1386-4564},
url = {https://doi.org/10.1007/s10791-018-9348-8},
journal = {Inf. Retr.},
month = {aug},
pages = {324–349},
numpages = {26},
}

@inproceedings{tie-aware-jcdl,
author = {Saha, Sourav and Roy, Dwaipayan and Mitra, Mandar},
title = {On modifying evaluation measures to deal with ties in ranked lists},
year = {2022},
isbn = {9781450393454},
publisher = {Association for Computing Machinery},
address = {New York, NY, USA},
url = {https://doi.org/10.1145/3529372.3533291},
doi = {10.1145/3529372.3533291},
booktitle = {Proceedings of the 22nd ACM/IEEE Joint Conference on Digital Libraries},
articleno = {12},
numpages = {4},
location = {Cologne, Germany},
series = {JCDL '22}
}
}

\end{document}